\documentclass[manuscript,screen]{acmart}

\usepackage{subcaption}
\usepackage{placeins}
\usepackage{float}
\usepackage{xcolor}
\usepackage{booktabs}
\usepackage[ruled,vlined]{algorithm2e}
\SetKwInput{Require}{Require}
\SetKwInput{Ensure}{Ensure}
\usepackage{multirow}

\AtBeginDocument{%
  }
\acmJournal{HEALTH}

\setcopyright{none}

\begin{document}
\title{Reliable Multilingual Orthopedic Decision Support from Clinical Narratives: Language-Aware Adaptation and Verification-Guided Deferral}

\author{Danish Ali}
\email{danish.ali@whu.edu.cn}
\orcid{0000-0000-0000-0000}
\affiliation{%
  \institution{School of Computer Science, Wuhan University}
  \city{Wuhan}
  \state{Hubei}
  \country{China}
}

\author{Li Xiaojian}
\authornote{Corresponding Author}
\email{lixiaojian@whu.edu.cn}
\orcid{0000-0000-0000-0000}
\affiliation{%
  \institution{School of Computer Science, Wuhan University}
  \city{Wuhan}
  \state{Hubei}
  \country{China}
}

\author{Sundas Iqbal}
\email{sundasiqbal058@gmail.com}
\orcid{0000-0000-0000-0000}
\affiliation{%
  \institution{School of Software, Nanjing University of Information Science and Technology (NUIST)}
  \city{Nanjing}
  \country{China}
}

\author{Farrukh Zaidi}
\email{farrukhzaidi001@gmail.com}
\affiliation{%
  \institution{Department of Orthopedic, Bahawal Victoria Hospital}
  \city{Bahawalpur, Punjab}
  \country{Pakistan}
}

\renewcommand{\shortauthors}{Ali et al.}

\begin{abstract}
Multilingual orthopedic decision support remains challenging in low-resource healthcare settings, where clinical narratives contain specialized terminology, mixed scripts, incomplete evidence, label imbalance and language-dependent documentation patterns. This article presents a reliability-oriented framework for classifying free-text orthopedic notes in English, Hindi and Punjabi. We compare task-aligned multilingual transformer encoders, a task-fine-tuned DistilBERT baseline, zero-shot instruction-tuned large language models (LLMs) and a domain-adaptive encoder, IndicBERT-HPA. IndicBERT-HPA augments IndicBERT with language-aware orthopedic adapter heads to support clinically relevant multilingual representation learning. Evaluation extends beyond aggregate accuracy to per-class performance, ROC-AUC, AUPRC, expected calibration error, cross-language stability and robustness under controlled balanced and natural-prevalence distributions. The evaluated zero-shot LLMs remain substantially less effective than task-adapted encoders for closed-set classification, with language-dependent instability. Under natural clinical prevalence, IndicBERT-HPA achieves the strongest overall performance, reaching an averaged Macro-F1 of 0.8792, Macro-AUROC of 0.894 and AUPRC of 0.902. We further implement a deterministic selective-verification layer combining confidence gating, evidence-consistency checking and language-risk screening. On a randomly selected held-out 5,000-record subset, it achieves 84.4\% selective accuracy and 0.76 selective Macro-F1 at 72.3\% coverage, compared with 71.5\% accuracy and 0.65 Macro-F1 for accept-all prediction. These results support reliability-oriented multilingual clinical decision support with explicit deferral.
\end{abstract}



\maketitle

\section{Introduction}
\label{sec:introduction}

Multilingual clinical decision support remains challenging in linguistically diverse healthcare environments \cite{abirami2026nlp}. In South Asian healthcare contexts, clinical interactions frequently occur in regional languages such as Hindi and Punjabi, while electronic health records and AI-based decision-support tools remain predominantly English-centric \cite{baker2025diagnostic,soni2025effective}. This language mismatch can limit equitable access to healthcare computing systems and contribute to inconsistent interpretation and care delivery \cite{garcia2025language}. Orthopedic decision support provides a structured and safety-sensitive setting for studying multilingual clinical narratives because patient descriptions often contain overlapping symptoms, incomplete evidence, heterogeneous terminology and language-dependent documentation patterns \cite{gao2021limitations,del2025exploring}. These challenges are amplified in low-resource languages, where annotated clinical data are limited and records may contain mixed scripts, non-standard terminology and variable documentation quality \cite{qiu2024towards,singh2024indicgenbench}. Recent multilingual healthcare benchmarking has further shown that clinical language-model performance can vary substantially across languages in low- and middle-income settings, reinforcing the need for language-conditioned evaluation \cite{restrepo2025multiophthalingua}. 

From a computing-for-healthcare perspective, multilingual orthopedic decision support requires reliable interpretation of noisy free-text clinical narratives under label skew, linguistic heterogeneity and evidential uncertainty \cite{villena2025nlp, qiu2024towards}. In a retrospective assistive setting, usefulness depends not only on predictive performance, but also on calibrated confidence, consistent behavior across languages and prevalence patterns and explicit deferral of uncertain cases for clinician review \cite{jain2025multilingual,he2021mdeberta}. Aggregate metrics may obscure clinically meaningful failure modes, including class-conditional confusion, language-conditional instability and overconfident uncertain predictions \cite{ji2024unified}. Consequently, multilingual orthopedic narrative classification should be studied as a reliability-oriented clinical decision-support task rather than as autonomous diagnosis or conventional text classification alone \cite{kim2025domain}.

Existing approaches can be grouped into task-only fine-tuning of multilingual encoders, supervised classification baselines and domain-adaptive architectures. Models such as IndicBERT, XLM-RoBERTa and mDeBERTa provide transferable multilingual representations \cite{kakwani2020indicnlpsuite,conneau2020xlmr}, but their behavior in safety-sensitive multilingual orthopedic decision support remains insufficiently characterized \cite{griot-etal-2025-pattern}. Furthermore, zero-shot instruction-tuned large language models (LLMs), despite strong generative fluency, may lack task-specific grounding and calibrated confidence for closed-set clinical categorization \cite{griot2025metacognition}\cite{zou2025uncertainty}. These limitations are especially important under natural clinical prevalence rather than balanced benchmark conditions. Accordingly, the present study addresses four linked challenges: orthopedic domain specialization, language-conditional instability, calibration under prevalence differences and deterministic post-prediction reliability control.

To address these challenges, we propose IndicBERT-HPA, a domain-adaptive multilingual encoder for orthopedic decision-support classification from clinical narratives in English, Hindi and Punjabi. The reported experiments use a curated multilingual orthopedic corpus derived from de-identified records collected at a tertiary-care hospital (anonymous) in Pakistan with collaboration of Indian hospitals. The corpus supports evaluation under both controlled balanced and natural-prevalence distributions. IndicBERT-HPA augments IndicBERT \cite{kakwani2020indicnlpsuite} with language-aware orthopedic adapter heads that project shared multilingual representations into a clinically discriminative space. We compare the proposed architecture with task-fine-tuned transformer baselines and zero-shot instruction-tuned LLMs \cite{zhang2025automedeval}.

Beyond model comparison, we implement and evaluate a deterministic selective-verification layer combining confidence gating, symptom--category evidence checking and language-risk screening. This layer does not replace the classifier, generate an alternative clinical category or modify model parameters; rather, it determines whether a prediction can be automatically accepted or should be deferred for human review. The main contributions are as follows:

\begin{enumerate}
    \item \textbf{Curated Multilingual Orthopedic Clinical Narrative Corpus.} 
    We construct a large-scale multilingual orthopedic clinical corpus spanning English, Hindi and Punjabi. The corpus supports both class-balanced and natural-prevalence evaluation, enabling analysis under class imbalance and language heterogeneity.

    \item \textbf{Domain-Adaptive Multilingual Encoder.} 
    We introduce IndicBERT-HPA, which augments IndicBERT with language-aware orthopedic adapter heads while preserving a shared multilingual representation space.

    \item \textbf{Reliability-Oriented Evaluation Protocol.} 
    We evaluate models using per-class performance, ROC-AUC, AUPRC, expected calibration error (ECE), cross-language stability and comparison across controlled balanced and natural-prevalence distributions.

    \item \textbf{Deterministic Selective-Verification Layer.} 
    We implement and retrospectively evaluate a post-prediction layer that combines confidence, evidence consistency and language-risk signals to authorize acceptance or defer uncertain predictions for review. Prospective clinician-in-the-loop validation remains future work.
\end{enumerate}

This study addresses the following research questions:

\begin{itemize}
    \item \textbf{RQ1:} How do language-conditioned natural-prevalence distributions affect the performance and calibration of multilingual orthopedic decision-support models relative to a controlled balanced setting?
    \item \textbf{RQ2:} Does explicit domain adaptation improve multilingual orthopedic decision-support classification performance and calibration compared with task-only fine-tuning?
    \item \textbf{RQ3:} How do task-fine-tuned encoders and zero-shot LLMs differ in classification performance and language-dependent failure modes for closed-set multilingual clinical categorization?
    \item \textbf{RQ4:} Can deterministic selective verification improve safety-oriented decision support by reducing automatically accepted errors and routing uncertain cases for human review?
\end{itemize}

The remainder of this paper is organized as follows. Section~\ref{sec:relatedwork} reviews related work. Section~\ref{sec:preliminaries} defines the task setting, notation and evaluation protocol. Section~\ref{sec:methodology} describes the dataset, proposed architecture, LLM evaluation and selective-verification layer. Section~\ref{sec:experiments} reports the experimental results. Sections~\ref{sec:discussion} and~\ref{sec:conclusion} discuss implications, limitations and future work.

\section{Related Work}
\label{sec:relatedwork}

This work intersects multilingual clinical NLP, clinical applications of LLMs, low-resource healthcare language processing and reliability-oriented clinical AI. We review these areas and position our contribution at their intersection.

\subsection{Multilingual Clinical NLP with Task-Aligned Encoders}

Transformer-based multilingual clinical NLP must account for linguistic variation, specialized terminology and limited annotated data \cite{qiu2024towards,raithel2025cross}. Multilingual encoders such as mBERT \cite{devlin2019bert}, XLM-RoBERTa \cite{conneau2020xlmr} and related architectures support cross-lingual representation learning, while biomedical benchmarking studies indicate that model behavior can vary substantially across tasks and clinical settings \cite{chen2025benchmarking}. Raithel et al. \cite{raithel2025cross} examined cross- and multilingual medication detection using transformer encoders and reported language-dependent variation. Villena et al. \cite{villena2025nlp} further emphasized careful modeling choices under restricted clinical data availability. Together, these studies motivate task-aligned and domain-aware multilingual modeling with explicit reliability evaluation \cite{qiu2024towards}.

\subsection{Large Language Models in Clinical Decision Support}

LLMs have increasingly been explored for diagnosis, triage and clinical decision support \cite{gaber2025llmcds,chen2025omnirag}. However, their suitability for structured safety-sensitive prediction requires careful evaluation \cite{posada2024evaluation,rose2025meddxagent}. Wu et al. \cite{wu2025large} evaluated LLM-assisted diagnostic performance in complex critical illness cases, highlighting the need to distinguish fluent responses from dependable clinical decisions. In orthopedics, Baker et al. \cite{baker2025diagnostic} assessed ChatGPT-4 for orthopedic oncology diagnosis. Bonfigli et al. \cite{bonfigli2024pre} similarly emphasized the importance of task adaptation and validation when deploying LLMs in biomedical contexts. In contrast to open-ended generation, our work evaluates zero-shot LLMs for fixed-label multilingual orthopedic classification.

\subsection{Low-Resource and Indic Language Healthcare NLP}

Multilingual healthcare NLP remains comparatively underexplored for low-resource languages \cite{thompson2019relevant,ling2025domain}. Indic-language modeling has benefited from pretrained resources such as IndicBERT \cite{kakwani2020indicnlpsuite}. Nazir et al. \cite{nazir2025leveraging} studied multilingual transformers for low-resource Indic-language classification, although not in a clinical diagnosis setting. Garcia-Lopez et al. \cite{garcia2025language} documented the impact of language barriers in musculoskeletal care, motivating multilingual orthopedic decision-support research \cite{soni2025effective}. Posada et al. \cite{posada2024evaluation} further reported limitations of zero-shot language-model configurations in resource-constrained medical text classification. Our study extends this direction through naturally authored orthopedic narratives in English, Hindi and Punjabi.

\subsection{Safety, Calibration and System-Level Design in Clinical AI}

Beyond predictive accuracy, clinical decision-support systems must support safety, interpretability and human accountability \cite{strong2025guideddeferral,ferdaus2026towards}. Musen et al. \cite{musen2021clinical} frame clinical decision-support systems as assistive rather than autonomous tools requiring human oversight. Recent methodological work has highlighted metric-related pitfalls in medical AI evaluation, including the importance of calibration, per-class analysis and threshold-independent assessment \cite{reinke2024metricpitfalls}, while clinical-text research further emphasizes aligning automated evaluation with real clinical workflows \cite{gan2025clinicalcoding,chaturvedi2025temporal}. Ethical and legal analyses similarly emphasize auditable separation between algorithmic prediction, clinical validation and decision authority \cite{nogaroli2025ethicalai}. Building on these principles, our deterministic selective-verification layer separates diagnostic prediction from authorization for automatic acceptance and evaluates the resulting accuracy--coverage trade-off in a multilingual clinical setting \cite{li2026clicare,corbeil2025modular}. 

\section{Preliminaries}
\label{sec:preliminaries}

This section formalizes the multilingual clinical text mining task, the data distributions used in the study, the diagnostic prediction setting, and the selective verification quantities used for reliability-oriented evaluation. The goal is to establish a common notation for subsequent sections. Architectural details of the proposed IndicBERT-HPA model and the deterministic verification layer are presented in Section~\ref{sec:methodology}.

\subsection{Multilingual Clinical Text Mining Setting}

We study multilingual orthopedic text classification as a structured knowledge discovery problem over noisy clinical narratives. Each clinical record consists of a free-text orthopedic complaint, a language identifier and a structured diagnostic label. Let
\begin{equation}
\mathcal{D}=\{(x_i,\ell_i,y_i)\}_{i=1}^{N}
\label{eq:dataset_definition}
\end{equation}
denote a multilingual clinical corpus, where $x_i$ is the free-text clinical note, $\ell_i$ is the language of the note, and $y_i$ is the corresponding diagnostic category. The language space is defined as
\begin{equation}
\mathcal{L}=\{\mathrm{EN},\mathrm{HI},\mathrm{PA}\},
\label{eq:language_space}
\end{equation}
where EN, HI, and PA denote English, Hindi, and Punjabi, respectively. The diagnostic label space is defined as
\begin{equation}
\mathcal{Y} =
\{\text{Spinal},\;
\text{Musculoskeletal},\;
\text{Bone},\;
\text{Hip},\;
\text{Other},\;
\text{Unknown}\}.
\label{eq:diag_label_space}
\end{equation}

The \emph{Unknown} category is part of the diagnostic label space and denotes insufficient or ambiguous diagnostic evidence in the input text. It should be distinguished from the verification-layer \emph{deferral decision}, which is a post-prediction action that routes a case to review rather than automatically accepting the model output.

\subsection{Controlled and Natural Data Distributions}

To evaluate both controlled model behavior and deployment-oriented robustness, we consider two empirical data distributions: a controlled balanced distribution and a natural-prevalence distribution. Let
\begin{equation}
\mathcal{D}^{\mathrm{bal}}=\{(x_i,\ell_i,y_i)\}_{i=1}^{N_{\mathrm{bal}}}
\label{eq:balanced_dataset}
\end{equation}
denote the controlled balanced dataset, where diagnostic classes are sampled approximately uniformly within each language. For each language $\ell \in \mathcal{L}$ and class $c \in \mathcal{Y}$, the class count is
\begin{equation}
n_{\ell,c}^{\mathrm{bal}}=
\sum_{i=1}^{N_{\mathrm{bal}}}
\mathbf{1}\{\ell_i=\ell,\;y_i=c\},
\label{eq:balanced_count}
\end{equation}
with $n_{\ell,c}^{\mathrm{bal}}$ kept approximately constant across classes. This setting enables controlled comparison of model behavior under uniform class exposure.

The natural-prevalence dataset is denoted as
\begin{equation}
\mathcal{D}^{\mathrm{nat}}=\{(x_i,\ell_i,y_i)\}_{i=1}^{N_{\mathrm{nat}}},
\label{eq:natural_dataset}
\end{equation}
where the empirical class distribution reflects realistic clinical prevalence. For each language $\ell$, the empirical class prior under distribution $d \in \{\mathrm{bal},\mathrm{nat}\}$ is
\begin{equation}
\pi_{\ell,c}^{d}
=
\frac{
\sum_{i=1}^{N_d}\mathbf{1}\{\ell_i=\ell,\;y_i=c\}
}{
\sum_{i=1}^{N_d}\mathbf{1}\{\ell_i=\ell\}
}.
\label{eq:class_prior}
\end{equation}
The difference between controlled and natural distributions captures label skew and prevalence shift. We quantify the language-conditioned distributional shift as
\begin{equation}
\Delta_{\ell}
=
\frac{1}{2}
\sum_{c\in\mathcal{Y}}
\left|
\pi_{\ell,c}^{\mathrm{nat}}
-
\pi_{\ell,c}^{\mathrm{bal}}
\right|.
\label{eq:distribution_shift}
\end{equation}
A larger $\Delta_{\ell}$ indicates stronger deviation between balanced evaluation and natural clinical prevalence for language $\ell$.

\subsection{Diagnostic Prediction Function}

Given an input pair $(x_i,\ell_i)$, a diagnostic model $f_{\theta}$ produces a probability distribution over the diagnostic label space:
\begin{equation}
\mathbf{p}_{\theta}(x_i,\ell_i)
=
\left[
p_{\theta}(y=1\mid x_i,\ell_i),
\dots,
p_{\theta}(y=|\mathcal{Y}|\mid x_i,\ell_i)
\right].
\label{eq:prob_output}
\end{equation}
The predicted diagnostic label is obtained by maximum posterior decision:
\begin{equation}
\hat{y}_i
=
\arg\max_{c\in\mathcal{Y}}
p_{\theta}(y=c\mid x_i,\ell_i),
\label{eq:prediction}
\end{equation}
and the associated confidence score is defined as
\begin{equation}
s_i
=
\max_{c\in\mathcal{Y}}
p_{\theta}(y=c\mid x_i,\ell_i).
\label{eq:confidence}
\end{equation}

In multilingual clinical settings, the confidence score is not only a ranking signal but also a reliability signal. Overconfident predictions under incomplete evidence, language mismatch or distribution shift may lead to unsafe automatic acceptance. Therefore, subsequent evaluation considers both discrimination and calibration.

\subsection{Language-Conditioned Performance}

For each language $\ell \in \mathcal{L}$, we define the language-specific subset as
\begin{equation}
\mathcal{D}_{\ell}^{d}
=
\{(x_i,\ell_i,y_i)\in\mathcal{D}^{d}\;|\;\ell_i=\ell\},
\quad
d\in\{\mathrm{bal},\mathrm{nat}\}.
\label{eq:language_subset}
\end{equation}
Model performance is computed separately for each $\mathcal{D}_{\ell}^{d}$ to expose language-conditioned instability that may be hidden by global averages. For class $c$, precision, recall, and F1-score are computed as
\begin{equation}
\mathrm{Precision}_{c}
=
\frac{\mathrm{TP}_{c}}
{\mathrm{TP}_{c}+\mathrm{FP}_{c}},
\quad
\mathrm{Recall}_{c}
=
\frac{\mathrm{TP}_{c}}
{\mathrm{TP}_{c}+\mathrm{FN}_{c}},
\label{eq:precision_recall}
\end{equation}
\begin{equation}
\mathrm{F1}_{c}
=
\frac{
2\cdot \mathrm{Precision}_{c}\cdot \mathrm{Recall}_{c}
}{
\mathrm{Precision}_{c}+\mathrm{Recall}_{c}
}.
\label{eq:f1_class}
\end{equation}
The macro-averaged F1-score is then defined as
\begin{equation}
\mathrm{MacroF1}
=
\frac{1}{|\mathcal{Y}|}
\sum_{c\in\mathcal{Y}}
\mathrm{F1}_{c}.
\label{eq:macro_f1}
\end{equation}
Macro-averaging is important in this study because natural clinical prevalence may be skewed, and minority diagnostic categories should not be dominated by frequent classes.

\subsection{Calibration and Ranking-Based Evaluation}
\label{subsec:calibration_ranking}

To evaluate discrimination independent of a fixed decision threshold, we report class-specific ROC-AUC, aggregate Macro-AUROC and AUPRC. ROC-AUC measures the separability of one diagnostic category from the remaining categories across false-positive and true-positive trade-offs. For each model and language partition, the aggregate Macro-AUROC reported in the summary tables is computed as the unweighted mean of the one-vs-rest ROC-AUC values across the six diagnostic categories:
\begin{equation}
\mathrm{Macro\text{-}AUROC}_{\ell}
=
\frac{1}{|\mathcal{Y}|}
\sum_{c \in \mathcal{Y}}
\mathrm{ROC\text{-}AUC}_{\ell,c},
\qquad |\mathcal{Y}| = 6,
\label{eq:macro_auroc}
\end{equation}
where $\ell \in \{\mathrm{EN}, \mathrm{HI}, \mathrm{PA}\}$ denotes the language partition and $c$ denotes a diagnostic category. Class-specific ROC-AUC values are reported in the per-category analysis, whereas Macro-AUROC is used in the language-wise and averaged summary tables. When results are summarized across languages, the reported value is the arithmetic mean of the language-specific Macro-AUROC scores:
\begin{equation}
\overline{\mathrm{Macro\text{-}AUROC}}
=
\frac{1}{|\mathcal{L}|}
\sum_{\ell \in \mathcal{L}}
\mathrm{Macro\text{-}AUROC}_{\ell},
\qquad |\mathcal{L}| = 3.
\label{eq:avg_macro_auroc}
\end{equation}
AUPRC is additionally reported because precision--recall behavior is informative under class imbalance, particularly in the natural-prevalence setting where minority diagnostic categories may otherwise be obscured by frequent classes.

Calibration is measured using expected calibration error (ECE). Let the prediction confidence interval $[0,1]$ be partitioned into $M$ bins $\{B_m\}_{m=1}^{M}$. For each bin $B_m$, let $\mathrm{acc}(B_m)$ denote empirical accuracy and $\mathrm{conf}(B_m)$ denote average confidence. ECE is defined as
\begin{equation}
\mathrm{ECE}
=
\sum_{m=1}^{M}
\frac{|B_m|}{N}
\left|
\mathrm{acc}(B_m)
-
\mathrm{conf}(B_m)
\right|.
\label{eq:ece}
\end{equation}
A lower ECE indicates better alignment between predicted confidence and empirical correctness. In safety-critical multilingual text mining, calibration is essential because selective verification depends on whether confidence scores meaningfully reflect prediction reliability.

\subsection{Selective Verification and Deferral}

The deterministic verification layer operates after the base classifier has produced $\hat{y}_i$ and $s_i$. It does not modify the model parameters. Instead, it decides whether the prediction should be automatically accepted or deferred for review. Let
\begin{equation}
g_{\phi}(x_i,\ell_i,\hat{y}_i,s_i)
\in
\{\mathrm{accept},\mathrm{defer}\}
\label{eq:verification_function}
\end{equation}
denote the verification decision function, where $\phi$ represents fixed verification rules and thresholds.

The verification layer combines three signals: confidence, symptom--diagnosis evidence consistency and language-consistency risk. Let
\begin{equation}
e_i = E(x_i,\hat{y}_i)
\in
\{\mathrm{supported},\mathrm{insufficient},\mathrm{contradicted}\}
\label{eq:evidence_signal}
\end{equation}
denote the evidence-consistency signal, and let
\begin{equation}
r_i = R(x_i,\ell_i)
\in
\{\mathrm{low},\mathrm{high}\}
\label{eq:language_risk}
\end{equation}
denote the language-consistency risk signal. Given a confidence threshold $\tau$, a conservative verification rule can be written as
\begin{equation}
g_{\phi}(x_i,\ell_i,\hat{y}_i,s_i)
=
\begin{cases}
\mathrm{defer}, & s_i < \tau \;\vee\; e_i \neq \mathrm{supported} \;\vee\; r_i=\mathrm{high},\\
\mathrm{accept}, & \text{otherwise}.
\end{cases}
\label{eq:deferral_rule}
\end{equation}
This rule formalizes the separation between prediction generation and decision authorization. The classifier proposes a diagnostic label, while the verification layer determines whether the prediction is reliable enough for automatic acceptance.

\subsection{Coverage, Selective Accuracy and Risk Reduction}

Selective prediction evaluates the system only over automatically accepted cases while also accounting for the proportion of cases deferred for review. Let
\begin{equation}
a_i = \mathbf{1}\{g_{\phi}(x_i,\ell_i,\hat{y}_i,s_i)=\mathrm{accept}\}
\label{eq:accept_indicator}
\end{equation}
be the automatic acceptance indicator. The coverage of the system is
\begin{equation}
\mathrm{Coverage}
=
\frac{1}{N}
\sum_{i=1}^{N} a_i,
\label{eq:coverage}
\end{equation}
and the deferral rate is
\begin{equation}
\mathrm{DeferralRate}
=
1-\mathrm{Coverage}.
\label{eq:deferral_rate}
\end{equation}
Selective accuracy measures accuracy among accepted predictions:
\begin{equation}
\mathrm{SelectiveAccuracy}
=
\frac{
\sum_{i=1}^{N}
a_i\mathbf{1}\{\hat{y}_i=y_i\}
}{
\sum_{i=1}^{N}a_i
}.
\label{eq:selective_accuracy}
\end{equation}
The accepted error count is defined as
\begin{equation}
U(g_{\phi})
=
\sum_{i=1}^{N}
a_i\mathbf{1}\{\hat{y}_i\neq y_i\}.
\label{eq:accepted_errors}
\end{equation}
Let $g_{\mathrm{all}}$ denote the baseline policy that accepts all predictions. The unsafe accepted error reduction is computed as
\begin{equation}
\mathrm{RiskReduction}
=
\frac{
U(g_{\mathrm{all}})-U(g_{\phi})
}{
U(g_{\mathrm{all}})
}.
\label{eq:risk_reduction}
\end{equation}
This quantity measures how much the verification layer reduces automatically accepted errors relative to an unconstrained prediction-only system.

\subsection{Evaluation Protocol}

Models are evaluated across English, Hindi and Punjabi under both controlled balanced and natural-prevalence distributions. Predictive performance is reported using accuracy, precision, recall, Macro-F1, class-specific ROC-AUC, aggregate Macro-AUROC and AUPRC. Calibration behavior is measured using ECE. Reliability under multilingual distribution shift is examined through language-conditioned performance and comparison between $\mathcal{D}^{\mathrm{bal}}$ and $\mathcal{D}^{\mathrm{nat}}$. For the selective verification layer, we additionally report coverage, deferral rate, selective accuracy, accepted error count and unsafe accepted error reduction. Unless otherwise specified, results are reported as mean $\pm$ standard deviation.

\section{Methodology}
\label{sec:methodology}

The methodology is explicitly aligned with the research questions and is designed to disentangle three commonly conflated paradigms in multilingual clinical NLP: task-only fine-tuning of general-purpose encoders, explicit domain-adaptive modeling and zero-shot generative reasoning with large language models. Our goal is not only to compare predictive performance but to understand which modeling choices yield reliable, interpretable and safety-compatible behavior in multilingual orthopedic decision support. 

\begin{figure}[h]
  \centering
  \includegraphics[width=0.9\linewidth]{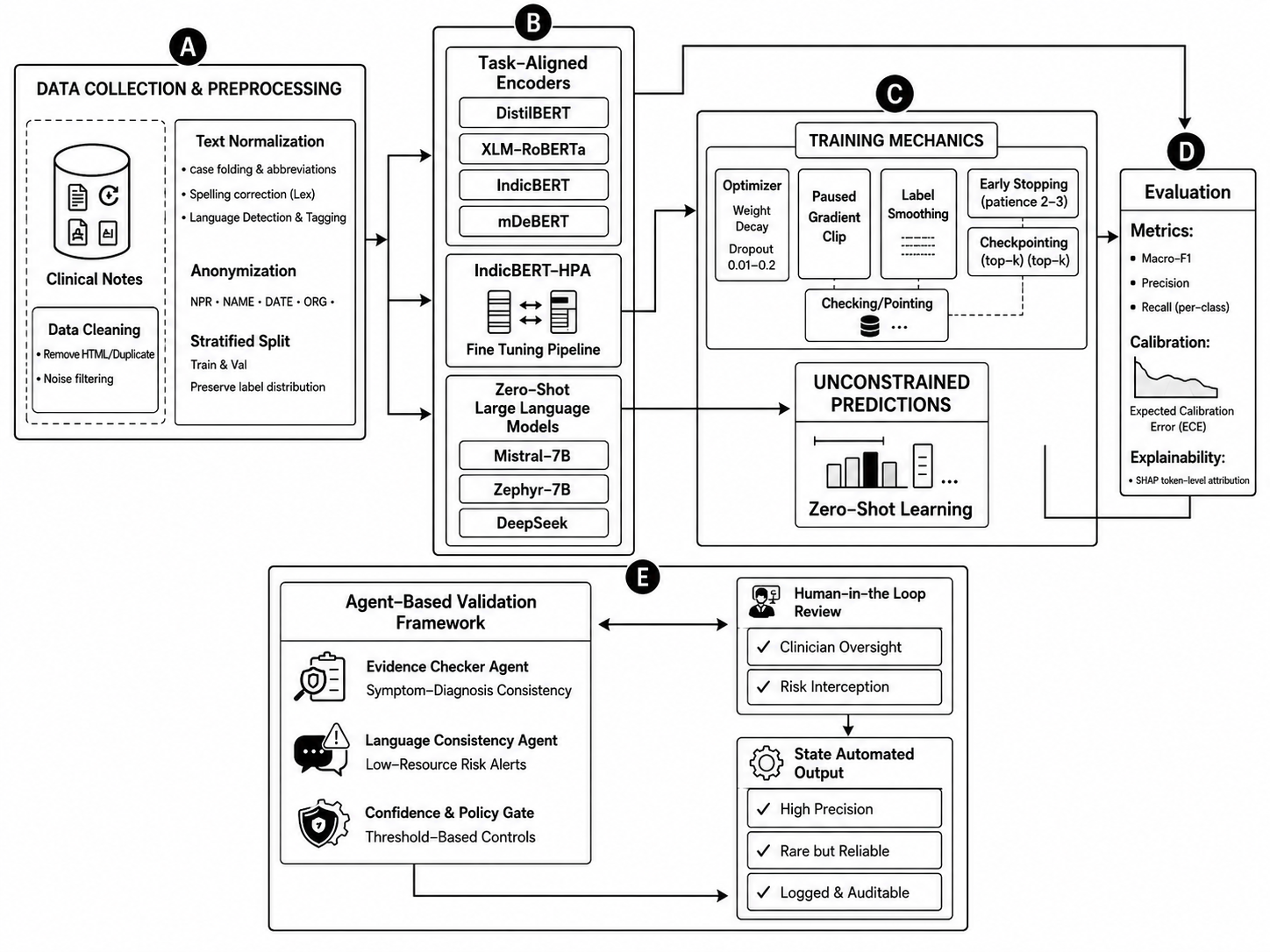}
  \caption{Overview of the proposed multilingual orthopedic decision-support framework. The pipeline integrates task-aligned transformer encoders, domain-adaptive representation learning, zero-shot large language model evaluation and a deterministic selective-verification layer for evidence-aware review deferral.}
  \Description{A block diagram illustrating the full experimental pipeline. Clinical text inputs undergo preprocessing and tokenization before being processed by task-fine-tuned transformer encoders and a domain-adaptive architecture. In parallel, zero-shot large language models generate diagnostic predictions. Outputs from all models are passed to an selective verification layer that performs evidence checking, language consistency analysis and escalation to human review when uncertainty or risk is detected}
  \label{fig:method_overview}
\end{figure}

Fig.~\ref{fig:method_overview} provides an end-to-end overview of the proposed experimental pipeline. The framework begins with multilingual clinical text preprocessing and proceeds through task-aligned encoder training, domain-adaptive representation learning and zero-shot large language model evaluation. These model outputs are subsequently integrated into a safety-oriented agent framework that enforces evidence validation and human-in-the-loop escalation. This staged design enables controlled analysis of representation learning, domain specialization and system-level reliability within a unified experimental setting. Accordingly, the methodology proceeds in four stages. First, we evaluate task-fine-tuned multilingual transformer encoders on a controlled orthopedic diagnosis benchmark with emphasis on low-resource languages (Hindi and Punjabi). Second, we analyze the failure modes of instruction-tuned large language models when applied in a zero-shot manner to structured clinical diagnosis. Third, we introduce and evaluate a domain-adaptive architecture (IndicBERT-HPA) that incorporates orthopedic specialization beyond task-only learning. Finally, building on these empirical findings, we design a deterministic selective-verification framework that augments model predictions with explicit evidence checks, language consistency analysis and conservative human-in-the-loop gating. Together, these components enable a system-level evaluation that reflects real-world clinical safety requirements rather than benchmark optimization alone.

\subsection{Problem Definition and Data Description}
\label{subsec:problem_data}

This work targets reliable clinical decision support for orthopedic diagnosis in multilingual, low-resource settings. In many Indian healthcare contexts, preliminary assessment and referral decisions are informed by free-text clinical notes recorded in regional languages, creating a practical gap for decision-support systems that are predominantly English-centric and domain-agnostic. We formulate the task as multilingual orthopedic diagnosis classification from unstructured clinical text. Given a clinical description $x$ written in English (EN), Hindi (HI) or Punjabi (PA), the system predicts a structured diagnostic category as defined in Eq.~\ref{eq:label_space}
\begin{equation}
y \in \mathcal{Y} = \{\text{spinal}, \text{musculoskeletal}, \text{bone}, \text{hip}, \text{other}, \text{unknown}\}
\label{eq:label_space}
\end{equation}
The goal is not autonomous diagnosis but a decision-support function that provides consistent preliminary categorization while explicitly exposing uncertainty for clinician oversight.

\vspace{0.3em}
\noindent

In terms of methodological challenges the task lies at the intersection of domain specificity and language resource constraints. Orthopedic narratives contain subtle symptom descriptions, anatomical references and context-dependent terminology that general-purpose language models are not optimized to interpret reliably. These challenges are amplified for Hindi and Punjabi, where clinical documentation frequently exhibits non-standard terminology, transliteration and mixed-script usage and where labeled medical data is limited. As a result, raw predictive performance alone is insufficient; evaluation must support failure analysis and safety-oriented system design under realistic multilingual conditions. Recent multilingual healthcare benchmarking has further shown that clinical LLM performance can vary substantially across languages in low- and middle-income settings, reinforcing the need for language-conditioned evaluation \cite{restrepo2025multiophthalingua}.

\subsection{Dataset Construction and Distribution Design}
\label{subsec:dataset}

The clinical records analyzed in this study were sourced from a tertiary-care hospital (anonymous) in Pakistan with collaboration of Indian hospitals. Although hospital (anonymous) maintains orthopedic collaboration with institutions in India, the dataset used in the present experiments was derived from the hospital (anonymous) clinical corpus. The corpus contains naturally authored orthopedic clinical notes in English, Hindi, and Punjabi; the Hindi and Punjabi notes were not generated through translation from English. Data curation and clinical label validation were conducted under the supervision of an orthopedic physician team. The source corpus contained more than $60{,}000$ naturally occurring orthopedic clinical notes per language. After removal of duplicate and low-information records, clinical label harmonization, de-identification and text normalization, $134{,}979$ records were retained for natural-prevalence evaluation.

The shared diagnostic label space is
\[
\mathcal{Y}=
\{\textit{spinal},\textit{musculoskeletal},\textit{bone},
\textit{hip},\textit{other},\textit{unknown}\}.
\]
We construct two complementary evaluation settings. The \textbf{controlled setting} contains $18{,}000$ records, obtained by selecting $1{,}000$ records from each diagnostic category within each language partition, yielding $6{,}000$ balanced records per language. The \textbf{natural-prevalence setting} retains the refined corpus with its observed class distribution. Table~\ref{tab:dataset_summary} summarizes both settings and Table~\ref{tab:natural_distribution} reports the natural-prevalence class distribution.

\begin{table}[t]
\centering
\caption{Summary of the multilingual orthopedic diagnosis dataset under controlled and natural-prevalence settings.}
\label{tab:dataset_summary}
\footnotesize
\setlength{\tabcolsep}{2pt}
\renewcommand{\arraystretch}{.80}
\begin{tabular}{lcccccc}
\toprule
\textbf{Setting} & \textbf{Lang.} & \textbf{Script} & \textbf{Records} &
\textbf{Classes} & \textbf{Per Class} & \textbf{Dist.} \\
\midrule
\multirow{3}{*}{Controlled}
& EN & Latin      & 6{,}000  & 6 & 1{,}000 & Balanced \\
& HI & Devanagari & 6{,}000  & 6 & 1{,}000 & Balanced \\
& PA & Gurmukhi   & 6{,}000  & 6 & 1{,}000 & Balanced \\
\midrule
\multirow{3}{*}{Natural}
& EN & Latin      & 44{,}979 & 6 & Variable & Skewed \\
& HI & Devanagari & 45{,}000 & 6 & Variable & Skewed \\
& PA & Gurmukhi   & 45{,}000 & 6 & Variable & Skewed \\
\midrule
\textbf{Total} & -- & -- & 134{,}979 & 6 & -- & Skewed \\
\bottomrule
\end{tabular}
\end{table}

\begin{table}[t]
\centering
\caption{Class distribution of the natural-prevalence dataset; values are count (percentage).}
\label{tab:natural_distribution}
\footnotesize
\setlength{\tabcolsep}{2pt}
\renewcommand{\arraystretch}{.80}
\begin{tabular}{lrrrr}
\toprule
\textbf{Class} & \textbf{English} & \textbf{Hindi} & \textbf{Punjabi} & \textbf{Overall} \\
\midrule
Spinal &
1{,}284 (2.85) &
1{,}176 (2.61) &
25{,}393 (56.43) &
27{,}853 (20.64) \\
Musculoskeletal &
8{,}236 (18.31) &
8{,}236 (18.30) &
6{,}116 (13.59) &
22{,}588 (16.73) \\
Bone &
6{,}112 (13.59) &
6{,}133 (13.63) &
6{,}112 (13.58) &
18{,}357 (13.60) \\
Hip &
24{,}245 (53.90) &
24{,}267 (53.93) &
2{,}120 (4.71) &
50{,}632 (37.51) \\
Other &
4{,}013 (8.92) &
4{,}051 (9.00) &
4{,}013 (8.92) &
12{,}077 (8.95) \\
Unknown &
1{,}089 (2.42) &
1{,}137 (2.53) &
1{,}246 (2.77) &
3{,}472 (2.57) \\
\midrule
\textbf{Total} &
\textbf{44{,}979 (100.00)} &
\textbf{45{,}000 (100.00)} &
\textbf{45{,}000 (100.00)} &
\textbf{134{,}979 (100.00)} \\
\bottomrule
\end{tabular}
\end{table}

The natural-prevalence setting exhibits substantial language-conditioned label variation: hip-related complaints dominate the English and Hindi partitions, whereas spinal complaints dominate the Punjabi partition. Therefore, the controlled setting supports balanced cross-language model comparison, while the natural-prevalence setting evaluates classification and calibration behavior under the observed retrospective hospital (anonymous) clinical distribution. Because class prevalence differs substantially across language partitions, language-conditioned performance should be interpreted jointly with the distributional characteristics reported in Table~\ref{tab:natural_distribution}.

\subsection{Data Preprocessing and Label Refinement}
\label{Preprocessing}

Clinical label refinement was conducted by an orthopedic physician team, who reviewed symptom descriptions against the predefined six-category diagnostic taxonomy to resolve ambiguous or inconsistent label assignments and maintain clinically meaningful category boundaries across English, Hindi and Punjabi records. The clinical notes were originally authored in their respective languages; the Hindi and Punjabi records were not generated through translation from English. During preprocessing, language-specific clinical expressions were preserved while duplicate records, low-information entries and patient-identifying information were removed. Consequently, the experimental corpus contains only de-identified symptom narratives, language identifiers and diagnostic-category labels required for the retrospective analyses reported in this study.

\subsection{Multilingual Transformer Models}
\label{subsec:encoders}

To establish supervised baselines for multilingual orthopedic diagnosis, we evaluate four transformer-based encoder models spanning English-centric and multilingual pretraining models. Each model is paired with a standard classification head to enable multi-class prediction over diagnostic categories.

\begin{itemize}

\item \textbf{XLM-RoBERTa}  \cite{conneau2020xlmr} :
A widely used multilingual transformer pretrained on large-scale cross-lingual corpora. It represents a strong general-purpose multilingual baseline and is commonly employed for cross-lingual text classification tasks.

\item \textbf{IndicBERT} \cite{kakwani2020indicnlpsuite}: 
A transformer model pretrained specifically on Indic languages. It serves as a language-aligned baseline for Hindi and Punjabi clinical text, allowing us to assess the benefits of Indic-centric pretraining.

\item \textbf{mDeBERTa} \cite{he2021mdeberta}: 
A multilingual variant of DeBERTa that employs disentangled attention mechanisms. This model is included to evaluate whether increased architectural complexity improves performance in low-resource clinical settings.

\item \textbf{DistilBERT} \cite{sanh2019distilbert}:  
DistilBERT is a lightweight English-centric transformer obtained via knowledge distillation from BERT. In this work, DistilBERT is used for the orthopedic diagnosis task by augmenting the pretrained encoder with a linear classification head. All encoder layers and the task head are updated using supervised clinical data, enabling us to examine how far task-level fine-tuning alone can compensate for linguistic and domain mismatch.
\end{itemize}

\subsection{Domain-Adaptive Architecture: IndicBERT-HPA}
\label{subsec:hpa}

To address domain and language mismatch in multilingual orthopedic text classification, we propose IndicBERT-HPA (Hindi--Punjabi Adapters), a domain-adaptive extension of IndicBERT with lightweight orthopedic adapter modules. The model uses a shared IndicBERT encoder for all languages and applies language-specific adapter transformations for Hindi and Punjabi. English follows the shared encoder path directly. This design keeps the main multilingual representation space shared while allowing targeted specialization for Hindi and Punjabi clinical expressions.

\paragraph{Architecture Overview.}
IndicBERT-HPA builds upon a shared IndicBERT encoder that produces contextual token representations. Let an input clinical sequence $x_i = (x_{i1}, \dots, x_{iT})$ with language tag $\ell_i \in \{\mathrm{EN}, \mathrm{HI}, \mathrm{PA}\}$ be encoded by IndicBERT as:
\begin{equation}
H_i = \mathrm{IndicBERT}(x_i) \in \mathbb{R}^{T \times d},
\label{eq:indicbert_encoder}
\end{equation}
where $d=768$ denotes the hidden dimensionality. We use the contextual representation of the \texttt{[CLS]} token as the sentence-level representation:
\begin{equation}
h_i = H_{i,\mathrm{CLS}} \in \mathbb{R}^{768}.
\label{eq:cls_representation}
\end{equation}

\paragraph{Language-Aware Adapter Routing.}
For Hindi and Punjabi inputs, IndicBERT-HPA applies a language-specific orthopedic adapter before classification. Each adapter is implemented as a bottleneck residual transformation:
\begin{equation}
A_{\ell_i}(h_i)
=
W^{(\ell_i)}_{2}
\,\sigma\!\left(
W^{(\ell_i)}_{1}h_i+b^{(\ell_i)}_{1}
\right)
+b^{(\ell_i)}_{2},
\label{eq:adapter_function}
\end{equation}
where $W^{(\ell_i)}_{1} \in \mathbb{R}^{r \times 768}$ and
$W^{(\ell_i)}_{2} \in \mathbb{R}^{768 \times r}$ are adapter weights,
$r=512$ is the bottleneck dimension, and $\sigma(\cdot)$ denotes a ReLU activation. The adapted representation is then defined as:
\begin{equation}
\tilde{h}_i =
\begin{cases}
h_i + A_{\ell_i}(h_i), & \ell_i \in \{\mathrm{HI}, \mathrm{PA}\},\\
h_i, & \ell_i = \mathrm{EN}.
\end{cases}
\label{eq:adapter_routing}
\end{equation}

Thus, Hindi and Punjabi each employ distinct orthopedic adapters, whereas English bypasses the adapter layer and uses the shared IndicBERT representation directly. This routing reflects the implemented architecture and avoids duplicating the full encoder while allowing targeted adaptation for low-resource multilingual clinical expressions.

\paragraph{Classification Objective.}
The final representation $\tilde{h}_i$ is passed to a linear classifier to produce probabilities over the $C=6$ orthopedic diagnostic categories:
\begin{equation}
p_i = p(y \mid x_i,\ell_i)
=
\mathrm{softmax}(W_c \tilde{h}_i + b_c),
\label{eq:classifier}
\end{equation}
where $W_c \in \mathbb{R}^{C \times 768}$ and $b_c \in \mathbb{R}^{C}$. The predicted label and confidence score are:
\begin{equation}
\hat{y}_i = \arg\max_{y} p_i(y),
\qquad
c_i = \max_{y} p_i(y).
\label{eq:hpa_prediction}
\end{equation}
The model is trained end-to-end using the cross-entropy objective:
\begin{equation}
\mathcal{L}
=
-\sum_{i=1}^{N}
\log p_i(y_i).
\label{eq:loss}
\end{equation}

\begin{figure}[h]
  \centering
  \includegraphics[width=0.7\linewidth]{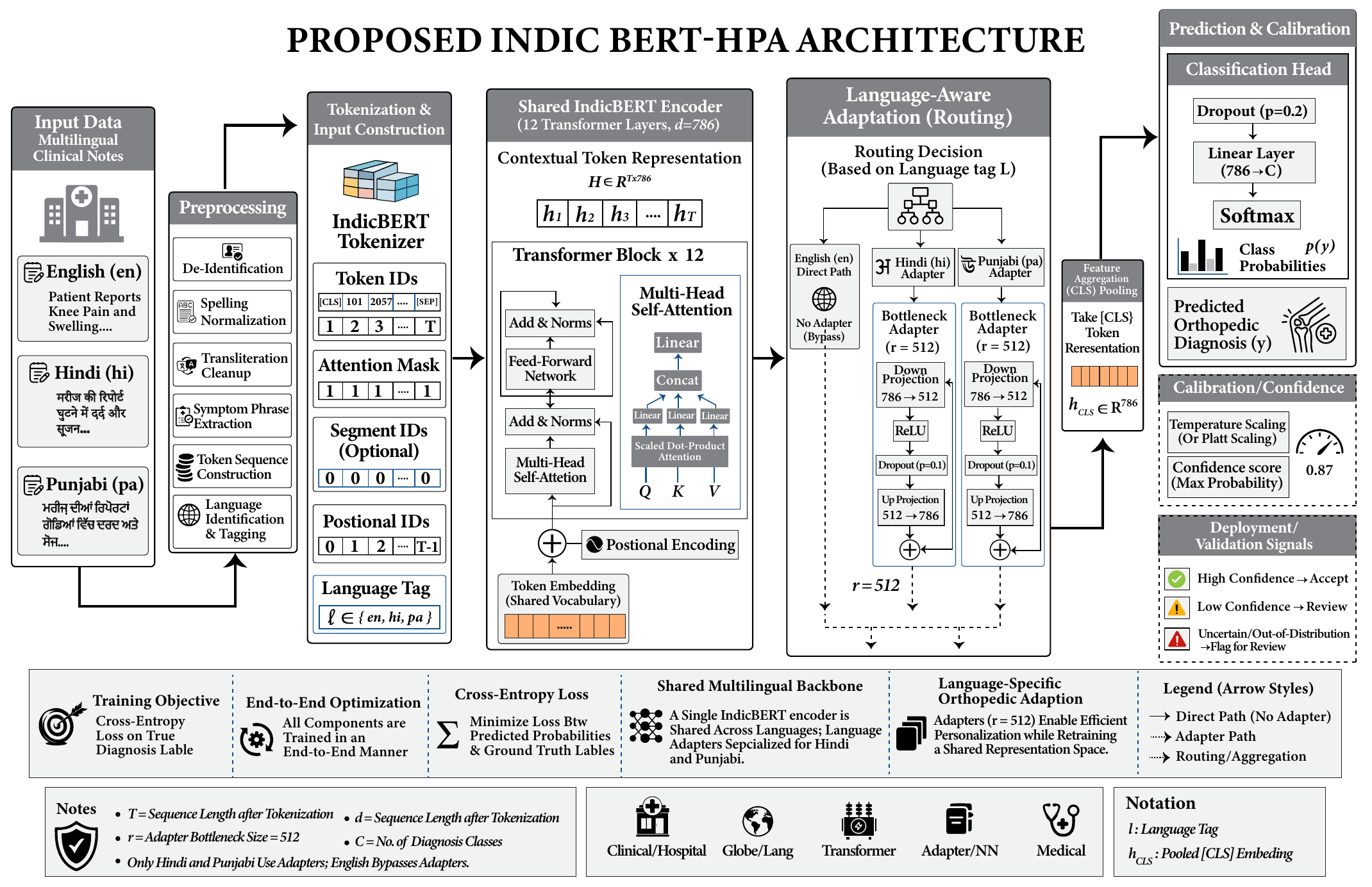}
  \caption{Proposed IndicBERT-HPA architecture. A shared IndicBERT encoder generates multilingual contextual representations. English follows the direct shared-encoder path, while Hindi and Punjabi representations are refined through language-specific orthopedic adapter heads before task-specific classification.}
  \label{fig:indicbert_hpa}
\end{figure}

\paragraph{Training Strategy and Rationale.}
During training, the shared IndicBERT encoder, Hindi and Punjabi adapter heads, and the classifier are jointly optimized. The adapter bottleneck constrains language-aware orthopedic specialization to a compact subspace, reducing the need to duplicate the full encoder while preserving shared multilingual representations. Compared with task-only fine-tuning, this design separates general multilingual encoding from targeted clinical-domain adaptation.

IndicBERT-HPA is based on the hypothesis that structured, language-aware domain adaptation can yield more reliable clinical predictions than supervised fine-tuning alone, particularly in low-resource and cross-lingual healthcare settings. As illustrated in Fig.~\ref{fig:indicbert_hpa}, the shared IndicBERT backbone first produces contextual representations. English inputs are classified directly from the shared representation, whereas Hindi and Punjabi inputs are routed through lightweight language-specific adapter modules that project representations into an orthopedic diagnostic subspace through residual bottleneck transformations. The resulting representation is then used for six-class orthopedic diagnosis. The complete training and diagnostic-prediction procedure for IndicBERT-HPA is summarized in Algorithm~\ref{alg:indicbert-hpa}. The algorithm makes the language-conditioned routing explicit: Hindi and Punjabi inputs are processed through their corresponding orthopedic adapter modules, whereas English inputs follow the direct shared-encoder path before classification.
\begin{algorithm}[t]
\scriptsize
\DontPrintSemicolon
\SetAlgoLined
\setlength{\algomargin}{0.7em}
\SetInd{0.25em}{0.6em}
\SetNlSkip{0pt}
\SetAlgoSkip{2pt}
\SetAlgoInsideSkip{2pt}
\caption{\textbf{IndicBERT-HPA Training and Diagnostic Prediction}}
\label{alg:indicbert-hpa}

\KwIn{
Training set $\mathcal{D}_{tr}=\{(x_i,y_i,\ell_i)\}$; validation set $\mathcal{D}_{va}$; \;
shared IndicBERT encoder $\mathcal{M}$ and tokenizer $\mathcal{T}$; \;
adapters $A_{\mathrm{HI}}$, $A_{\mathrm{PA}}$; classifier $(W_c,b_c)$.
}
\KwOut{
Trained parameters $\Theta=\{\mathcal{M},A_{\mathrm{HI}},A_{\mathrm{PA}},W_c,b_c\}$.
}

\BlankLine
\textbf{Training}\;
Initialize $A_{\mathrm{HI}}$, $A_{\mathrm{PA}}$, $W_c$, and $b_c$\;

\For{$e=1$ \KwTo $E$}{
  \ForEach{mini-batch $\{(x_i,y_i,\ell_i)\}_{i=1}^{B}\sim\mathcal{D}_{tr}$}{
    $\mathbf{t}_i \leftarrow \mathcal{T}(x_i)$;\;
    $\mathbf{H}_i \leftarrow \mathcal{M}(\mathbf{t}_i)$;\;
    $\mathbf{h}_i \leftarrow \mathbf{H}_{i,\mathrm{CLS}}$;\;

    \eIf{$\ell_i=\mathrm{HI}$}{
      $\tilde{\mathbf{h}}_i
      \leftarrow
      \mathbf{h}_i + A_{\mathrm{HI}}(\mathbf{h}_i)$\;
    }{
      \eIf{$\ell_i=\mathrm{PA}$}{
        $\tilde{\mathbf{h}}_i
        \leftarrow
        \mathbf{h}_i + A_{\mathrm{PA}}(\mathbf{h}_i)$\;
      }{
        $\tilde{\mathbf{h}}_i
        \leftarrow
        \mathbf{h}_i$
        \tcp*[f]{English direct path}\;
      }
    }

    $\mathbf{p}_i
    \leftarrow
    \mathrm{softmax}(W_c\tilde{\mathbf{h}}_i+b_c)$;\;
    $\mathcal{L}
    \leftarrow
    -\frac{1}{B}\sum_{i=1}^{B}\log \mathbf{p}_i[y_i]$;\;
    Update $\Theta$ using AdamW\;
  }
  Evaluate on $\mathcal{D}_{va}$ using Macro-F1 and ECE;\;
  Retain the best checkpoint
  \tcp*[f]{ECE used as tie-breaker}\;
}

\BlankLine
\textbf{Diagnostic Prediction}\;
\KwData{Clinical note $x$, language tag $\ell$}\;
$\mathbf{h}\leftarrow
\mathcal{M}(\mathcal{T}(x))_{\mathrm{CLS}}$;\;

\eIf{$\ell=\mathrm{HI}$}{
  $\tilde{\mathbf{h}}\leftarrow
  \mathbf{h}+A_{\mathrm{HI}}(\mathbf{h})$\;
}{
  \eIf{$\ell=\mathrm{PA}$}{
    $\tilde{\mathbf{h}}\leftarrow
    \mathbf{h}+A_{\mathrm{PA}}(\mathbf{h})$\;
  }{
    $\tilde{\mathbf{h}}\leftarrow\mathbf{h}$
    \tcp*[f]{English direct path}\;
  }
}

$\mathbf{p}\leftarrow
\mathrm{softmax}(W_c\tilde{\mathbf{h}}+b_c)$;\;
$\hat{y}\leftarrow
\arg\max_{y\in\mathcal{Y}}\mathbf{p}[y]$;\;
$c\leftarrow
\max_{y\in\mathcal{Y}}\mathbf{p}[y]$;\;

\Return $(\hat{y},c)$\;
\end{algorithm}

\subsection{Implementation Details and Reproducibility}
\label{subsec:implementation_details}

All supervised encoder models are trained under the same experimental protocol using stratified record-level train, validation and test partitions of 70\%/15\%/15\% for both the controlled and natural-prevalence settings, preserving language and diagnostic-label proportions. Models are implemented using the HuggingFace Transformers library and optimized with AdamW using a learning rate of $2\times10^{-5}$, weight decay of $0.01$, $\beta_1=0.9$, $\beta_2=0.999$, and $\epsilon=10^{-8}$. Clinical notes are truncated or padded to a maximum sequence length of 256 tokens. We use a batch size of 16, dropout of 0.20, gradient clipping with maximum norm 1.0, and a maximum of 10 epochs. Early stopping is applied with a patience of 3 epochs using validation Macro-F1, with ECE as a tie-breaker for checkpoint selection. IndicBERT-HPA uses an adapter bottleneck dimension of $r=512$ and follows the same training protocol as the encoder baselines. Experiments are conducted using random seeds $\{42,123,2025\}$, and repeated-run results are reported as mean $\pm$ standard deviation.

The validation set is also used to select the confidence threshold for the deterministic selective verification layer. We evaluate $\tau \in \{0.50,0.55,0.60,0.65,0.70,0.75\}$ and select the operating point that balances selective accuracy and deferral burden, yielding $\tau=0.60$. This threshold is fixed before test evaluation. No test-set information is used for checkpoint selection, threshold tuning, evidence-resource construction or rule calibration.

\subsection{Large Language Models and Zero-Shot Evaluation}
\label{subsec:llms}

In addition to supervised encoders, we evaluate three open instruction-tuned LLMs in a zero-shot closed-label setting: DeepSeek Open \cite{qiao2025deepseek}, Mistral-7B Instruct \cite{mistralai2023mistralinstruct} and Zephyr-7B \cite{huggingface2023zephyr}. The exact evaluated checkpoints are \nolinkurl{deepseek-ai/deepseek-llm-7b-chat}, \nolinkurl{mistralai/Mistral-7B-Instruct-v0.2}, and \nolinkurl{HuggingFaceH4/zephyr-7b-beta}, respectively. For each clinical note, the model is required to select exactly one category from the same six-label orthopedic taxonomy used by the supervised encoders. No demonstrations, retrieval augmentation, task-specific fine-tuning or parameter updates are used. This comparison therefore evaluates zero-shot closed-set behavior rather than task-adapted LLM performance.

\subsubsection{Prompting Protocol}
\label{subsubsec:llm_prompting}

All three LLMs are evaluated using the same zero-shot prompt template. The original clinical note is retained in its source language, while the output label set is fixed in English to maintain a unified decision space across English, Hindi and Punjabi inputs.

\begin{quote}
\footnotesize
\ttfamily
You are given an anonymized orthopedic clinical note. Classify the note into exactly one of the following diagnostic categories: Spinal, Musculoskeletal, Bone, Hip, Other, Unknown. Use Unknown only when the note does not contain enough evidence for any specific category. Return only a JSON object with two fields: label and confidence. The confidence must be a number between 0 and 1. Do not provide explanations. Clinical note: \{clinical\_note\}. Output:
\end{quote}

All models are evaluated using deterministic decoding with \texttt{do\_sample=False}, \texttt{max\_new\_tokens=120} and one generation attempt per record. Zephyr-7B is loaded using 4-bit NF4 quantization with \texttt{float16} computation.

\subsubsection{Label Mapping and Invalid Output Handling}
\label{subsubsec:llm_label_mapping}

Generated responses are parsed automatically and normalized to the predefined diagnostic label set before metric computation. Normalization removes superficial capitalization, punctuation, and whitespace variation while preserving the selected diagnostic category. A response is considered valid only when it can be mapped unambiguously to exactly one of the six predefined labels. Responses containing conflicting categories, labels outside the predefined taxonomy, or malformed outputs from which no single valid category can be recovered are counted as incorrect predictions rather than reassigned to \textit{unknown}. Accuracy, precision, recall and Macro-F1 are reported under the controlled and natural-prevalence settings using the same diagnostic label space as the supervised encoder evaluation.

\subsection{Deterministic Selective-Verification Layer}
\label{sec:verification}

We implement a deterministic selective-verification layer that separates diagnostic prediction from decision authorization. Given a prediction $\hat{y}_i$ and confidence score $c_i$ produced by IndicBERT-HPA for input $x_i$ in language $\ell_i$, the layer evaluates symptom--diagnosis evidence consistency and language-related risk before determining whether the prediction should be automatically accepted or deferred for review. The layer applies fixed, auditable verification rules and does not perform autonomous reasoning, generate an alternative diagnosis or modify the underlying classifier.

The verification layer consists of three deterministic components. The \textbf{Evidence-Consistency Checker} compares the predicted category with a physician-reviewed symptom--diagnosis resource and returns
$e_i \in \{\mathrm{supported}, \mathrm{insufficient}, \mathrm{contradicted}\}$.
The \textbf{Language-Risk Screener} checks script mismatch, abnormal script mixing, excessive transliteration noise and insufficient language-specific clinical evidence, returning
$r_i \in \{\mathrm{low}, \mathrm{high}\}$.
The \textbf{Deferral Authorization Gate} combines these signals with model confidence and returns an accept-or-defer decision without modifying the predicted diagnostic category.

The final verification decision is defined as:
\begin{equation}
g(x_i,\ell_i,\hat{y}_i,c_i)=
\begin{cases}
\mathrm{defer}, & c_i < \tau \;\lor\; e_i \neq \mathrm{supported} \;\lor\; r_i=\mathrm{high},\\
\mathrm{accept}, & \mathrm{otherwise},
\end{cases}
\label{eq:safety_rule}
\end{equation}
where $\tau=0.60$ is selected on the validation set and fixed before test evaluation. Thus, low-confidence, weakly supported, contradictory or language-risk predictions are deferred rather than forced into automatic acceptance. Importantly, \textit{unknown} remains a diagnostic label, whereas \textit{defer} is a post-prediction authorization decision.

Algorithm~\ref{alg:selective-verification} summarizes the implemented procedure.

\begin{algorithm}[t]
\scriptsize
\DontPrintSemicolon
\SetAlgoLined
\setlength{\algomargin}{0.7em}
\SetInd{0.25em}{0.6em}
\SetNlSkip{0pt}
\SetAlgoSkip{2pt}
\SetAlgoInsideSkip{2pt}
\caption{\textbf{Deterministic Selective Verification}}
\label{alg:selective-verification}

\KwIn{
Clinical note $x$; language tag $\ell$; predicted label $\hat{y}$; \;
confidence $c$; fixed threshold $\tau=0.60$.
}
\KwOut{
Authorization decision $g\in\{\mathrm{accept},\mathrm{defer}\}$.
}

$e\leftarrow\textsc{EvidenceCheck}(x,\hat{y})$
\tcp*[f]{supported / insufficient / contradicted}\;

$r\leftarrow\textsc{LanguageRisk}(x,\ell)$
\tcp*[f]{low / high}\;

\eIf{$c<\tau$ \textbf{or} $e\neq\mathrm{supported}$ \textbf{or} $r=\mathrm{high}$}{
  $g\leftarrow\mathrm{defer}$\;
}{
  $g\leftarrow\mathrm{accept}$\;
}

\Return $(\hat{y},c,g)$\;
\end{algorithm}

\begin{figure}[t]
\centering
\includegraphics[width=0.7\linewidth]{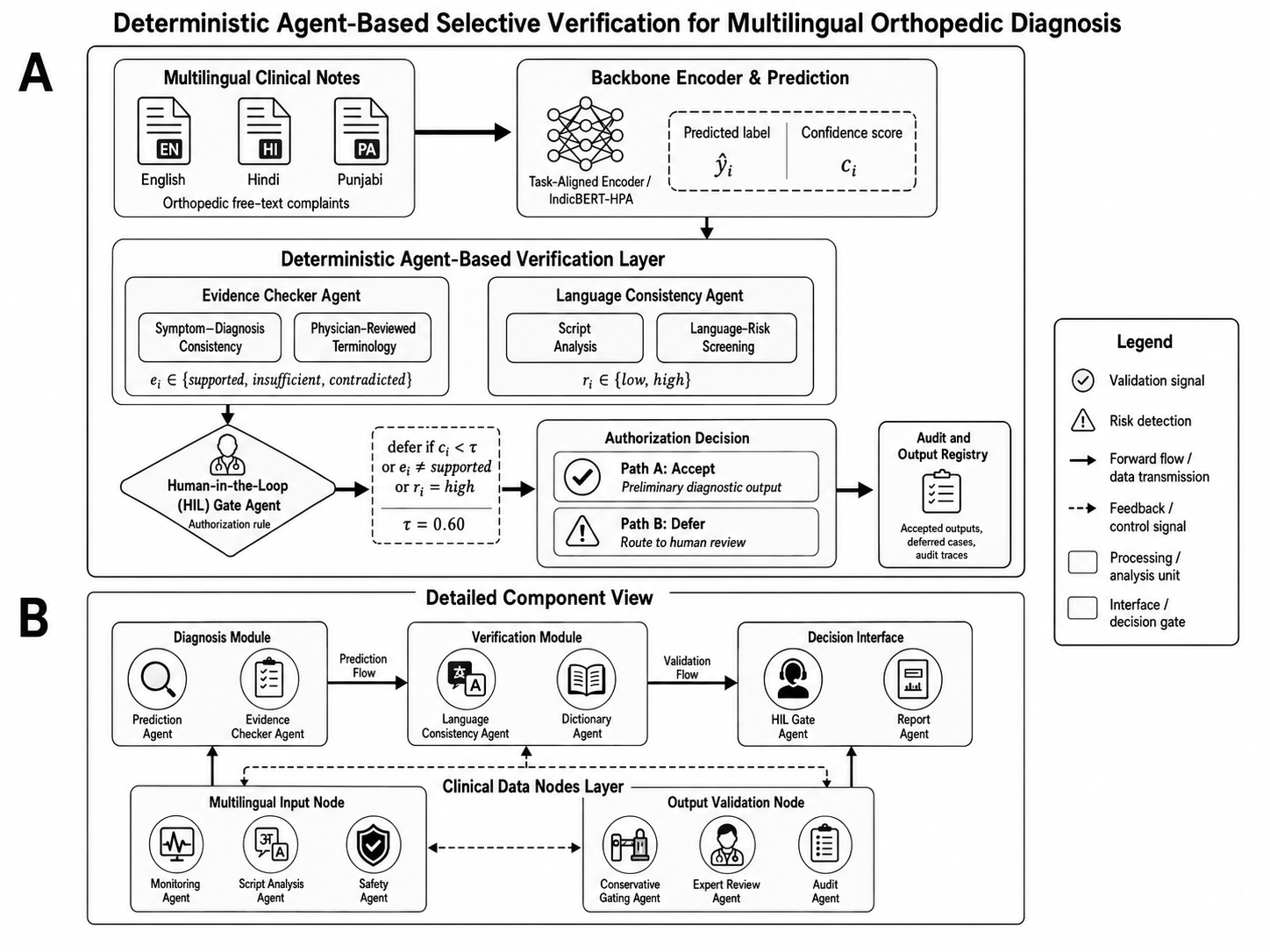}
\caption{Deterministic selective-verification layer combining confidence gating, symptom--diagnosis evidence checking, language-risk screening, and review deferral.}
\label{fig:verification_framework}
\end{figure}

The complete verification workflow is illustrated in Fig.~\ref{fig:verification_framework}. IndicBERT-HPA first produces a diagnostic hypothesis and confidence score; the deterministic verification components then determine whether the prediction is suitable for automatic acceptance or should be routed for review.

\subsubsection{Evidence Dictionary and Language-Risk Construction}
\label{subsubsec:evidence_language_rules}

The evidence resource is constructed from diagnostic label definitions, clinician-reviewed symptom descriptions, anatomical indicators, and language-specific terminology observed during training-set curation. Canonical cues are aligned across English, Hindi, and Punjabi and are used only for verification; test-set labels are not used for dictionary expansion or rule calibration. Language-risk rules use the expected scripts for English (Latin), Hindi (Devanagari), and Punjabi (Gurmukhi), marking records as high risk when they exhibit strong script mismatch, abnormal mixed-script usage, excessive transliteration noise, or insufficient language-specific clinical evidence.

\subsubsection{Operational Properties and Scope}
\label{subsubsec:verification_principles}

The layer is deterministic, auditable, and conservative: identical inputs produce identical authorization decisions, each deferral is attributable to explicit confidence, evidence, or language-risk signals, and uncertain predictions are routed for review rather than automatically accepted. The layer operates only on model outputs and fixed verification resources; it does not retrain the classifier, generate an alternative diagnosis, or constitute a deployed clinical workflow. Its evaluation in this study is retrospective and is intended to measure selective-prediction reliability under multilingual clinical conditions.
\section{Experimental Results and Analysis}
\label{sec:experiments}

This section evaluates multilingual orthopedic text classification under controlled and natural-prevalence settings. We compare supervised transformer encoders with zero-shot LLMs for six-class prediction across English, Hindi and Punjabi, and then evaluate the proposed deterministic selective verification layer as a post-prediction mechanism for accepting reliable outputs or deferring uncertain cases for review. The analysis focuses on classification performance, calibration behavior, language-specific variation and selective reliability.

We first report results under the controlled balanced setting in Section~\ref{cs}, which supports class-wise and cross-language comparison under uniform class exposure. Results under the natural-prevalence setting are then presented in Section~\ref{rcd} to examine model behavior under the retrospective clinical distribution and its associated class imbalance.

\subsection{Controlled Experimental Setting}
\label{cs}

\subsubsection{Per-Class Performance Across Languages}
\label{subsec:per_class_performance}

The controlled setting supports class-wise comparison under balanced exposure across languages and diagnostic categories. Tables~\ref{tab:spinal_disorders_c}--\ref{tab:unknown_c} report Precision, Recall, F1 and ROC-AUC for English (EN), Hindi (HI) and Punjabi (PA), presented in a fixed EN/HI/PA order within each cell. Since F1 is calculated from the reported precision and recall values, it is shown without a separate standard deviation.

\subsubsection{Spinal Disorders}

Spinal-disorder classification exhibits clear language-dependent variation at the fixed decision threshold. IndicBERT-HPA achieves the highest English F1 (0.873) and the highest or tied-highest ROC-AUC across English, Hindi and Punjabi (0.926, 0.953 and 0.938, respectively). XLM-RoBERTa attains the highest Hindi F1 (0.784), while DistilBERT obtains the highest Punjabi F1 (0.818). Overall, the results in Table~\ref{tab:spinal_disorders_c} indicate strong cross-language class separation by IndicBERT-HPA, although selected baselines remain competitive in threshold-dependent prediction.
\begin{table}[H]
 \caption{Prediction performance for \textbf{Spinal disorders} under the controlled setting.}
  \label{tab:spinal_disorders_c}
  \centering
  \renewcommand{\arraystretch}{.90}
  \resizebox{\columnwidth}{!}{%
  \begin{tabular}{lcccc}
    \toprule
    \textbf{Model} & \textbf{Precision $\pm$ Std} & \textbf{Recall $\pm$ Std} & \textbf{F1} & \textbf{ROC-AUC $\pm$ Std} \\
    \midrule
     & \textbf{EN / HI / PA} & \textbf{EN / HI / PA} & \textbf{EN / HI / PA} & \textbf{EN / HI / PA} \\
    \midrule
    DistilBERT &
    0.779 $\pm$ 0.030 / 0.712 $\pm$ 0.020 / \textbf{0.862 $\pm$ 0.020} &
    0.734 $\pm$ 0.030 / \textbf{0.856 $\pm$ 0.020} / 0.778 $\pm$ 0.030 &
    0.756 / 0.777 / \textbf{0.818} &
    0.913 $\pm$ 0.020 / 0.926 $\pm$ 0.010 / 0.904 $\pm$ 0.020 \\
    
    XLM-RoBERTa &
    0.672 $\pm$ 0.030 / 0.723 $\pm$ 0.030 / 0.784 $\pm$ 0.030 &
    0.668 $\pm$ 0.030 / \textbf{0.856 $\pm$ 0.020} / \textbf{0.789 $\pm$ 0.030} &
    0.670 / \textbf{0.784} / 0.786 &
    0.923 $\pm$ 0.020 / \textbf{0.953 $\pm$ 0.010} / 0.937 $\pm$ 0.020 \\
    
    IndicBERT &
    0.491 $\pm$ 0.050 / 0.657 $\pm$ 0.030 / 0.527 $\pm$ 0.050 &
    0.477 $\pm$ 0.050 / 0.720 $\pm$ 0.030 / 0.608 $\pm$ 0.030 &
    0.484 / 0.687 / 0.565 &
    0.486 $\pm$ 0.050 / 0.468 $\pm$ 0.050 / 0.472 $\pm$ 0.050 \\
    
    mDeBERTa &
    0.302 $\pm$ 0.050 / 0.349 $\pm$ 0.050 / 0.434 $\pm$ 0.050 &
    0.316 $\pm$ 0.050 / 0.432 $\pm$ 0.050 / 0.386 $\pm$ 0.050 &
    0.309 / 0.386 / 0.409 &
    0.494 $\pm$ 0.050 / 0.482 $\pm$ 0.050 / 0.460 $\pm$ 0.050 \\
    
    IndicBERT-HPA (Proposed) &
    \textbf{0.894 $\pm$ 0.020} / \textbf{0.738 $\pm$ 0.030} / 0.720 $\pm$ 0.030 &
    \textbf{0.853 $\pm$ 0.020} / 0.816 $\pm$ 0.020 / 0.783 $\pm$ 0.030 &
    \textbf{0.873} / 0.775 / 0.750 &
    \textbf{0.926 $\pm$ 0.020} / \textbf{0.953 $\pm$ 0.010} / \textbf{0.938 $\pm$ 0.020} \\
    
    \bottomrule
  \end{tabular}
  }
  
\end{table}

\subsubsection{Musculoskeletal Disorders}

Musculoskeletal disorders represent the most consistent category-level advantage for IndicBERT-HPA in the controlled setting. The proposed model achieves the highest F1-score in English, Hindi and Punjabi (0.798, 0.894 and 0.798), together with the highest ROC-AUC values (0.921, 0.947 and 0.926). XLM-RoBERTa remains competitive in Hindi with an F1-score of 0.883, whereas IndicBERT and mDeBERTa show weaker discrimination for this category. These comparisons are detailed in Table~\ref{tab:musculoskeletal_disorders_c}.

\begin{table}[H]
  \caption{Prediction performance for \textbf{Musculoskeletal disorders} under the controlled setting.}
  \label{tab:musculoskeletal_disorders_c}
  \centering
  \renewcommand{\arraystretch}{.90}
  \resizebox{\columnwidth}{!}{%
  \begin{tabular}{lcccc}
    \toprule
    \textbf{Model} & \textbf{Precision $\pm$ Std} & \textbf{Recall $\pm$ Std} & \textbf{F1} & \textbf{ROC-AUC $\pm$ Std} \\
    \midrule
     & \textbf{EN / HI / PA} & \textbf{EN / HI / PA} & \textbf{EN / HI / PA} & \textbf{EN / HI / PA} \\
    \midrule
    DistilBERT &
    0.631 $\pm$ 0.030 / 0.851 $\pm$ 0.020 / 0.783 $\pm$ 0.030 &
    0.673 $\pm$ 0.030 / 0.802 $\pm$ 0.020 / 0.725 $\pm$ 0.030 &
    0.651 / 0.826 / 0.753 &
    0.890 $\pm$ 0.020 / 0.920 $\pm$ 0.020 / 0.911 $\pm$ 0.010 \\

    XLM-RoBERTa &
    0.664 $\pm$ 0.030 / 0.883 $\pm$ 0.020 / 0.731 $\pm$ 0.030 &
    0.690 $\pm$ 0.030 / 0.883 $\pm$ 0.020 / 0.779 $\pm$ 0.030 &
    0.677 / 0.883 / 0.754 &
    0.878 $\pm$ 0.020 / 0.917 $\pm$ 0.020 / 0.909 $\pm$ 0.010 \\

    IndicBERT &
    0.441 $\pm$ 0.050 / 0.644 $\pm$ 0.030 / 0.617 $\pm$ 0.030 &
    0.549 $\pm$ 0.050 / 0.673 $\pm$ 0.030 / 0.629 $\pm$ 0.030 &
    0.489 / 0.658 / 0.623 &
    0.455 $\pm$ 0.050 / 0.459 $\pm$ 0.050 / 0.477 $\pm$ 0.050 \\

    mDeBERTa &
    0.315 $\pm$ 0.050 / 0.394 $\pm$ 0.050 / 0.378 $\pm$ 0.050 &
    0.285 $\pm$ 0.050 / 0.388 $\pm$ 0.050 / 0.377 $\pm$ 0.050 &
    0.299 / 0.391 / 0.377 &
    0.470 $\pm$ 0.050 / 0.493 $\pm$ 0.050 / 0.468 $\pm$ 0.050 \\

    IndicBERT-HPA (Proposed) &
    \textbf{0.777 $\pm$ 0.030} / \textbf{0.893 $\pm$ 0.020} / \textbf{0.799 $\pm$ 0.030} &
    \textbf{0.820 $\pm$ 0.020} / \textbf{0.896 $\pm$ 0.020} / \textbf{0.798 $\pm$ 0.030} &
    \textbf{0.798} / \textbf{0.894} / \textbf{0.798} &
    \textbf{0.921 $\pm$ 0.010} / \textbf{0.947 $\pm$ 0.020} / \textbf{0.926 $\pm$ 0.020} \\
    \bottomrule
  \end{tabular}
  }
\end{table}

\subsubsection{Bone-Related Disorders}

Bone-related classification reveals metric-specific competition among the leading encoders. IndicBERT-HPA attains the highest F1-score in English (0.810) and Hindi (0.896), together with the highest ROC-AUC in Hindi (0.941) and Punjabi (0.946). XLM-RoBERTa leads Punjabi F1-score (0.814), whereas DistilBERT achieves the highest English ROC-AUC (0.943). As summarized in Table~\ref{tab:bone-related_disorders_c}, IndicBERT-HPA provides the strongest overall profile for this category while selected baselines remain competitive in specific language--metric comparisons.

\begin{table}[H]
  \caption{Prediction performance for \textbf{Bone-related disorders} under the controlled setting.}
  \label{tab:bone-related_disorders_c}
  \centering
  \renewcommand{\arraystretch}{.90}
  \resizebox{\columnwidth}{!}{%
  \begin{tabular}{lcccc}
    \toprule
    \textbf{Model} &
    \textbf{Precision $\pm$ Std} &
    \textbf{Recall $\pm$ Std} &
    \textbf{F1} &
    \textbf{ROC-AUC $\pm$ Std} \\
    \midrule
     &
     \textbf{EN / HI / PA} &
     \textbf{EN / HI / PA} &
     \textbf{EN / HI / PA} &
     \textbf{EN / HI / PA} \\
    \midrule
    DistilBERT &
    0.660$\pm$0.030 / 0.804$\pm$0.020 / 0.750$\pm$0.030 &
    0.621$\pm$0.030 / 0.845$\pm$0.020 / 0.779$\pm$0.030 &
    0.640 / 0.824 / 0.764 &
    \textbf{0.943$\pm$0.020} / 0.914$\pm$0.020 / 0.931$\pm$0.010 \\

    XLM-RoBERTa &
    0.695$\pm$0.030 / 0.800$\pm$0.030 / 0.829$\pm$0.020 &
    0.655$\pm$0.030 / 0.819$\pm$0.020 / \textbf{0.800$\pm$0.030} &
    0.674 / 0.809 / \textbf{0.814} &
    0.930$\pm$0.020 / 0.935$\pm$0.020 / 0.906$\pm$0.020 \\

    IndicBERT &
    0.556$\pm$0.050 / 0.640$\pm$0.030 / 0.625$\pm$0.030 &
    0.531$\pm$0.050 / 0.623$\pm$0.030 / 0.585$\pm$0.050 &
    0.543 / 0.631 / 0.604 &
    0.475$\pm$0.050 / 0.478$\pm$0.050 / 0.472$\pm$0.050 \\

    mDeBERTa &
    0.342$\pm$0.050 / 0.382$\pm$0.050 / 0.376$\pm$0.050 &
    0.316$\pm$0.050 / 0.397$\pm$0.050 / 0.412$\pm$0.050 &
    0.328 / 0.389 / 0.393 &
    0.485$\pm$0.050 / 0.481$\pm$0.050 / 0.469$\pm$0.050 \\

    IndicBERT-HPA (Proposed) &
    \textbf{0.850$\pm$0.020} / \textbf{0.939$\pm$0.020} / \textbf{0.842$\pm$0.020} &
    \textbf{0.774$\pm$0.030} / \textbf{0.857$\pm$0.020} / 0.716$\pm$0.030 &
    \textbf{0.810} / \textbf{0.896} / 0.774 &
    0.895$\pm$0.020 / \textbf{0.941$\pm$0.020} / \textbf{0.946$\pm$0.020} \\
    \bottomrule
  \end{tabular}
  }
\end{table}

\subsubsection{Hip-Related Disorders}

Hip-related disorders show distinct language-specific precision--recall trade-offs among the leading models. IndicBERT-HPA achieves the highest English F1-score (0.801) and records the highest ROC-AUC in Hindi (0.944) and Punjabi (0.930). At the fixed classification threshold, XLM-RoBERTa obtains the highest Hindi F1-score (0.851), whereas DistilBERT leads in Punjabi F1-score (0.836) and English ROC-AUC (0.910). The comparisons in Table~\ref{tab:hip-related_disorders_c} indicate that IndicBERT-HPA offers strong cross-language class separation, although threshold-dependent performance remains sensitive to language-specific behavior.

\begin{table}[H]
  \caption{Prediction performance for \textbf{Hip-related disorders} under the controlled setting.}
  \label{tab:hip-related_disorders_c}
  \centering
  \renewcommand{\arraystretch}{.90}
  \resizebox{\columnwidth}{!}{%
  \begin{tabular}{lcccc}
    \toprule
    \textbf{Model} &
    \textbf{Precision $\pm$ Std} &
    \textbf{Recall $\pm$ Std} &
    \textbf{F1} &
    \textbf{ROC-AUC $\pm$ Std} \\
    \midrule
     &
     \textbf{EN / HI / PA} &
     \textbf{EN / HI / PA} &
     \textbf{EN / HI / PA} &
     \textbf{EN / HI / PA} \\
    \midrule

    DistilBERT &
    0.667$\pm$0.030 / 0.738$\pm$0.030 / 0.805$\pm$0.020 &
    0.716$\pm$0.030 / 0.851$\pm$0.020 / \textbf{0.870$\pm$0.020} &
    0.691 / 0.790 / \textbf{0.836} &
    \textbf{0.910$\pm$0.020} / 0.925$\pm$0.020 / 0.912$\pm$0.020 \\

    XLM-RoBERTa &
    0.572$\pm$0.050 / 0.818$\pm$0.020 / 0.727$\pm$0.030 &
    0.720$\pm$0.030 / \textbf{0.886$\pm$0.020} / 0.754$\pm$0.030 &
    0.638 / \textbf{0.851} / 0.740 &
    0.873$\pm$0.010 / 0.902$\pm$0.020 / 0.894$\pm$0.020 \\

    IndicBERT &
    0.557$\pm$0.050 / 0.770$\pm$0.030 / 0.556$\pm$0.050 &
    0.559$\pm$0.050 / 0.640$\pm$0.030 / 0.616$\pm$0.030 &
    0.558 / 0.699 / 0.584 &
    0.486$\pm$0.050 / 0.484$\pm$0.050 / 0.475$\pm$0.050 \\

    mDeBERTa &
    0.312$\pm$0.050 / 0.423$\pm$0.050 / 0.391$\pm$0.050 &
    0.314$\pm$0.050 / 0.399$\pm$0.050 / 0.388$\pm$0.050 &
    0.313 / 0.411 / 0.389 &
    0.471$\pm$0.050 / 0.468$\pm$0.050 / 0.462$\pm$0.050 \\

    IndicBERT-HPA (Proposed) &
    \textbf{0.848$\pm$0.020} / \textbf{0.832$\pm$0.020} / \textbf{0.902$\pm$0.020} &
    \textbf{0.759$\pm$0.030} / 0.857$\pm$0.020 / 0.684$\pm$0.030 &
    \textbf{0.801} / 0.844 / 0.778 &
    0.904$\pm$0.010 / \textbf{0.944$\pm$0.010} / \textbf{0.930$\pm$0.020} \\

    \bottomrule
  \end{tabular}%
  }
\end{table}

\subsubsection{Other}

The heterogeneous \textit{Other} category presents variable performance across languages and evaluation measures. IndicBERT-HPA achieves the highest F1-score in English (0.797) and Punjabi (0.861), while also attaining the highest ROC-AUC in Hindi (0.926) and Punjabi (0.944). For Hindi F1, DistilBERT and XLM-RoBERTa share the highest rounded score (0.802), whereas DistilBERT records the strongest English ROC-AUC (0.954). As reported in Table~\ref{tab:other_c}, the proposed model remains particularly effective for Punjabi and selected Hindi comparisons, although no single model dominates every language--metric combination in this heterogeneous category.

\begin{table}[H]
  \caption{Prediction performance for \textbf{Other} under the controlled setting.}
  \label{tab:other_c}
  \centering
  \renewcommand{\arraystretch}{.90}
  \resizebox{\columnwidth}{!}{%
  \begin{tabular}{lcccc}
    \toprule
    \textbf{Model} &
    \textbf{Precision $\pm$ Std} &
    \textbf{Recall $\pm$ Std} &
    \textbf{F1} &
    \textbf{ROC-AUC $\pm$ Std} \\
    \midrule
     &
     \textbf{EN / HI / PA} &
     \textbf{EN / HI / PA} &
     \textbf{EN / HI / PA} &
     \textbf{EN / HI / PA} \\
    \midrule

    DistilBERT &
    0.699$\pm$0.030 / 0.724$\pm$0.030 / 0.831$\pm$0.020 &
    0.711$\pm$0.030 / 0.898$\pm$0.020 / 0.773$\pm$0.030 &
    0.705 / \textbf{0.802} / 0.801 &
    \textbf{0.954$\pm$0.010} / 0.918$\pm$0.020 / 0.898$\pm$0.010 \\

    XLM-RoBERTa &
    0.683$\pm$0.030 / \textbf{0.734$\pm$0.030} / 0.731$\pm$0.030 &
    0.657$\pm$0.030 / 0.885$\pm$0.020 / 0.721$\pm$0.030 &
    0.670 / \textbf{0.802} / 0.726 &
    0.850$\pm$0.010 / 0.878$\pm$0.010 / 0.895$\pm$0.010 \\

    IndicBERT &
    0.518$\pm$0.050 / 0.604$\pm$0.030 / 0.564$\pm$0.050 &
    0.568$\pm$0.050 / 0.707$\pm$0.030 / 0.570$\pm$0.050 &
    0.542 / 0.651 / 0.567 &
    0.468$\pm$0.050 / 0.477$\pm$0.050 / 0.459$\pm$0.050 \\

    mDeBERTa &
    0.334$\pm$0.050 / 0.392$\pm$0.050 / 0.420$\pm$0.050 &
    0.339$\pm$0.050 / 0.373$\pm$0.050 / 0.377$\pm$0.050 &
    0.336 / 0.382 / 0.397 &
    0.461$\pm$0.050 / 0.469$\pm$0.050 / 0.480$\pm$0.050 \\

    IndicBERT-HPA (Proposed) &
    \textbf{0.779$\pm$0.030} / 0.680$\pm$0.030 / \textbf{0.925$\pm$0.020} &
    \textbf{0.815$\pm$0.020} / \textbf{0.915$\pm$0.020} / \textbf{0.806$\pm$0.020} &
    \textbf{0.797} / 0.780 / \textbf{0.861} &
    0.912$\pm$0.020 / \textbf{0.926$\pm$0.020} / \textbf{0.944$\pm$0.020} \\

    \bottomrule
  \end{tabular}
  }
\end{table}

\subsubsection{Unknown}

The \textit{Unknown} category is important for safety-oriented evaluation because it represents notes with insufficient or ambiguous evidence for a specific orthopedic category and remains distinct from the post-prediction \textit{defer} decision. IndicBERT-HPA achieves the highest F1-score in English, Hindi and Punjabi (0.768, 0.874 and 0.844), together with the highest ROC-AUC values (0.911, 0.939 and 0.941). The results in Table~\ref{tab:unknown_c} therefore indicate stronger recognition of insufficient-evidence cases across languages in the controlled setting.

\begin{table}[H]
  \caption{Prediction performance for \textbf{Unknown} under the controlled setting.}
  \label{tab:unknown_c}
  \centering
  \renewcommand{\arraystretch}{.90}
  \resizebox{\columnwidth}{!}{%
  \begin{tabular}{lcccc}
    \toprule
    \textbf{Model} &
    \textbf{Precision $\pm$ Std} &
    \textbf{Recall $\pm$ Std} &
    \textbf{F1} &
    \textbf{ROC-AUC $\pm$ Std} \\
    \midrule
     &
     \textbf{EN / HI / PA} &
     \textbf{EN / HI / PA} &
     \textbf{EN / HI / PA} &
     \textbf{EN / HI / PA} \\
    \midrule

    DistilBERT &
    \textbf{0.841$\pm$0.020} / 0.872$\pm$0.020 / 0.795$\pm$0.030 &
    0.639$\pm$0.030 / 0.807$\pm$0.020 / 0.764$\pm$0.030 &
    0.726 / 0.838 / 0.779 &
    0.883$\pm$0.020 / 0.902$\pm$0.020 / 0.917$\pm$0.010 \\

    XLM-RoBERTa &
    0.659$\pm$0.030 / 0.792$\pm$0.030 / 0.758$\pm$0.030 &
    0.755$\pm$0.030 / 0.781$\pm$0.020 / 0.826$\pm$0.020 &
    0.704 / 0.786 / 0.791 &
    0.842$\pm$0.010 / 0.871$\pm$0.010 / 0.898$\pm$0.020 \\

    IndicBERT &
    0.579$\pm$0.050 / 0.724$\pm$0.030 / 0.679$\pm$0.030 &
    0.497$\pm$0.050 / 0.636$\pm$0.030 / 0.538$\pm$0.050 &
    0.535 / 0.677 / 0.600 &
    0.469$\pm$0.050 / 0.485$\pm$0.050 / 0.474$\pm$0.050 \\

    mDeBERTa &
    0.298$\pm$0.050 / 0.396$\pm$0.050 / 0.376$\pm$0.050 &
    0.348$\pm$0.050 / 0.417$\pm$0.050 / 0.411$\pm$0.050 &
    0.321 / 0.406 / 0.393 &
    0.488$\pm$0.050 / 0.477$\pm$0.050 / 0.484$\pm$0.050 \\

    IndicBERT-HPA (Proposed) &
    0.704$\pm$0.030 / \textbf{0.873$\pm$0.020} / \textbf{0.851$\pm$0.010} &
    \textbf{0.845$\pm$0.020} / \textbf{0.876$\pm$0.020} / \textbf{0.838$\pm$0.020} &
    \textbf{0.768} / \textbf{0.874} / \textbf{0.844} &
    \textbf{0.911$\pm$0.010} / \textbf{0.939$\pm$0.010} / \textbf{0.941$\pm$0.010} \\

    \bottomrule
  \end{tabular}
  }
\end{table}

\paragraph{Class-wise findings.}
Across the six diagnostic categories, IndicBERT-HPA demonstrates the strongest overall controlled-setting profile. It consistently leads musculoskeletal and safety-relevant \textit{Unknown} classification across all three languages, while recording the highest or tied-highest ROC-AUC in most language--category comparisons. Selected baselines remain competitive at specific fixed-threshold operating points, particularly for spinal and hip disorders in Hindi and Punjabi. Overall, these findings show that stronger class separation does not always yield the highest threshold-dependent F1-score, motivating the deterministic selective-verification analysis presented later.

\subsection{Performance of Task-Aligned Transformer Models}
\label{subsec:controlled_encoder_results}

Table~\ref{tab:encoder_models} reports supervised encoder performance under the controlled balanced setting. IndicBERT-HPA achieves the highest F1-Macro and AUPRC in each language partition, reaching F1-Macro values of 0.8078, 0.8441, and 0.8011 for English, Hindi, and Punjabi, respectively. For Macro-AUROC, IndicBERT-HPA achieves the strongest performance in Hindi and Punjabi, whereas DistilBERT records a slightly higher value in English. In terms of accuracy, IndicBERT-HPA performs best in English and Hindi, while DistilBERT obtains the highest Punjabi accuracy. IndicBERT-HPA records higher ECE than several baselines, particularly in English, indicating that improved classification performance does not by itself ensure well-calibrated confidence estimates.

\begin{table}[H]
\centering
\caption{Task-aligned encoder performance by language under the controlled setting. Macro-AUROC denotes the unweighted mean of the one-vs-rest ROC-AUC values across the six diagnostic categories.}
\label{tab:encoder_models}

\scriptsize
\setlength{\tabcolsep}{2pt}
\renewcommand{\arraystretch}{0.80}

\begin{tabular}{llccccc}
\toprule
\textbf{Model} & \textbf{Lang.} & \textbf{F1-Macro} & \textbf{Macro-AUROC} & \textbf{AUPRC} & \textbf{ECE} & \textbf{Accuracy} \\
\midrule
DistilBERT & EN & 0.6948 & \textbf{0.9155} & 0.8290 & 0.0748 & 0.6823 \\
           & HI & 0.8096 & 0.9175 & 0.9098 & 0.0084 & 0.8432 \\
           & PA & 0.7919 & 0.9122 & 0.8651 & 0.0222 & \textbf{0.7815} \\
\midrule
IndicBERT  & EN & 0.5252 & 0.4732 & 0.5667 & 0.0034 & 0.5302 \\
           & HI & 0.6674 & 0.4752 & 0.6667 & \textbf{0.0028} & 0.6665 \\
           & PA & 0.5906 & 0.4715 & 0.5667 & 0.0034 & 0.5910 \\
\midrule
IndicBERT-HPA (Proposed)
           & EN & \textbf{0.8078} & 0.9115 & \textbf{0.9022} & 0.1593 & \textbf{0.8110} \\
           & HI & \textbf{0.8441} & \textbf{0.9417} & \textbf{0.9317} & 0.0957 & \textbf{0.8695} \\
           & PA & \textbf{0.8011} & \textbf{0.9375} & \textbf{0.9101} & 0.0815 & 0.7708 \\
\midrule
mDeBERTa   & EN & 0.3179 & 0.4782 & 0.1667 & 0.0091 & 0.3197 \\
           & HI & 0.3943 & 0.4783 & 0.2000 & 0.0066 & 0.4010 \\
           & PA & 0.3931 & 0.4705 & 0.1849 & 0.0072 & 0.3918 \\
\midrule
XLM-RoBERTa & EN & 0.6720 & 0.8827 & 0.7718 & \textbf{0.0023} & 0.6908 \\
            & HI & 0.8193 & 0.9093 & 0.8248 & 0.0059 & 0.8517 \\
            & PA & 0.7686 & 0.9065 & 0.8693 & \textbf{0.0020} & 0.7782 \\
\bottomrule
\end{tabular}
\end{table}


\begin{figure}[H]
  \centering
  \includegraphics[width=\linewidth]{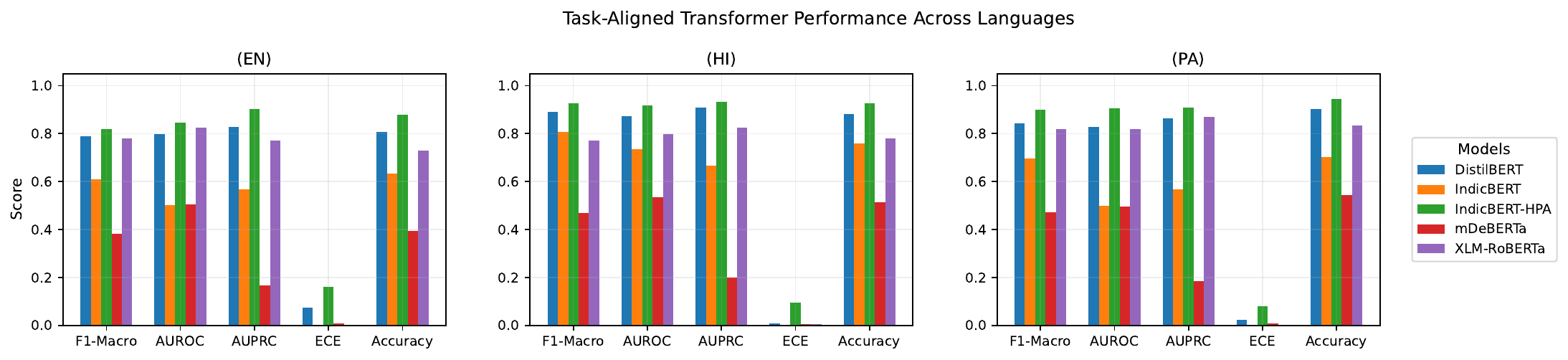}
  \caption{Controlled-setting performance of task-aligned encoders across English, Hindi, and Punjabi.}
  \label{fig:task_aligned_models}
\end{figure}

Averaged across languages, IndicBERT-HPA achieves the strongest overall controlled-setting profile, with 0.8177 F1-Macro, 0.9302 Macro-AUROC, 0.9147 AUPRC, and 0.8171 accuracy (Table~\ref{tab:average_results}). DistilBERT and XLM-RoBERTa remain competitive baselines, with XLM-RoBERTa obtaining slightly higher averaged accuracy than DistilBERT, whereas DistilBERT obtains higher averaged Macro-AUROC. Although IndicBERT obtains the lowest averaged ECE, its substantially lower F1-Macro and Macro-AUROC show that calibration error must be interpreted together with discriminative performance. The higher ECE of IndicBERT-HPA therefore motivates its use as the diagnostic backbone for the subsequent deterministic selective-verification analysis rather than direct automatic acceptance based on confidence alone.

\begin{table}[H]
\centering
\caption{Average encoder performance across languages under the controlled setting. Macro-AUROC is averaged across diagnostic categories and subsequently summarized across languages.}
\label{tab:average_results}

\scriptsize
\setlength{\tabcolsep}{2pt}
\renewcommand{\arraystretch}{0.80}

\begin{tabular}{lccccc}
\toprule
\textbf{Model} & \textbf{F1-Macro} & \textbf{Macro-AUROC} & \textbf{AUPRC} & \textbf{ECE} & \textbf{Accuracy} \\
\midrule
IndicBERT-HPA (Proposed) & \textbf{0.8177} & \textbf{0.9302} & \textbf{0.9147} & 0.1122 & \textbf{0.8171} \\
DistilBERT                & 0.7654 & 0.9151 & 0.8680 & 0.0351 & 0.7690 \\
XLM-RoBERTa               & 0.7533 & 0.8995 & 0.8220 & 0.0034 & 0.7736 \\
IndicBERT                 & 0.5944 & 0.4733 & 0.6000 & \textbf{0.0032} & 0.5959 \\
mDeBERTa                  & 0.3684 & 0.4757 & 0.1839 & 0.0076 & 0.3708 \\
\bottomrule
\end{tabular}
\end{table}


\begin{figure}[H]
  \centering
  \includegraphics[width=\linewidth]{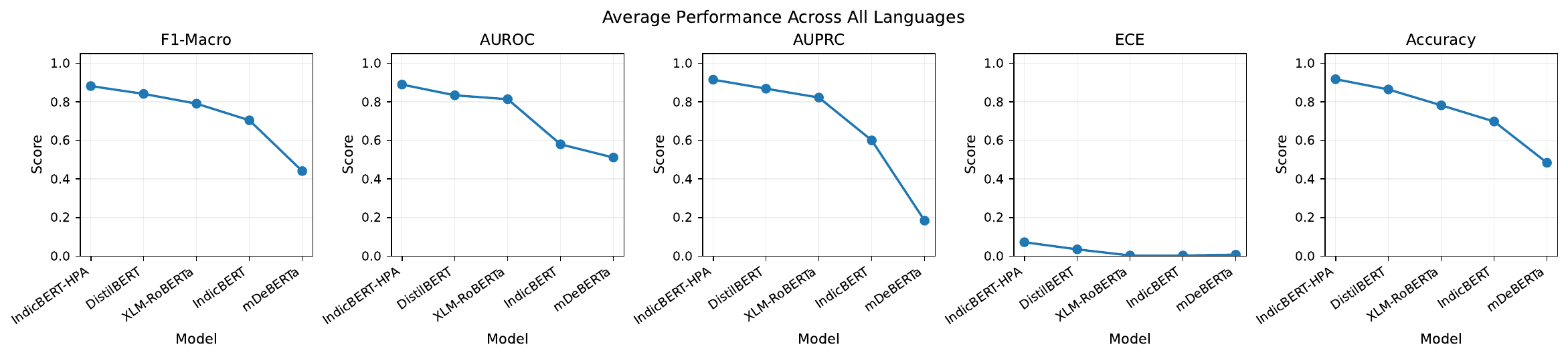}
  \caption{Average controlled-setting performance across languages; calibration differences motivate subsequent deterministic selective verification.}
  \label{fig:avg_small_multiples}
\end{figure}

\subsection{Zero-Shot Evaluation of Large Language Models}
\label{subsec:controlled_llm_results}

We evaluate the three zero-shot LLMs under the controlled multilingual setting using the protocol defined in Section~\ref{subsec:llms}. As shown in Table~\ref{tab:llm_models}, all evaluated LLMs perform substantially below the supervised encoder models for closed-set orthopedic classification. DeepSeek Open achieves the highest F1-score in English (0.2152) and Hindi (0.3279), whereas Mistral-7B Instruct achieves the highest Punjabi F1-score (0.1737). Overall, the results indicate limited zero-shot label grounding and considerable language-dependent variation in this structured clinical task.

\begin{table}[H]
\centering
\caption{Zero-shot LLM performance by language under the controlled setting.}
\label{tab:llm_models}

\scriptsize
\setlength{\tabcolsep}{2pt}
\renewcommand{\arraystretch}{0.80}

\begin{tabular}{llcccc}
\toprule
\textbf{Model} & \textbf{Lang.} & \textbf{Accuracy} & \textbf{Precision} & \textbf{Recall} & \textbf{F1-Score} \\
\midrule
DeepSeek Open & EN & 0.2560 & 0.1856 & 0.2560 & \textbf{0.2152} \\
              & HI & 0.3144 & 0.3481 & 0.3100 & \textbf{0.3279} \\
              & PA & 0.2231 & 0.0915 & 0.2200 & 0.1292 \\
\midrule
Mistral-7B Instruct & EN & 0.2740 & 0.1639 & 0.2740 & 0.2051 \\
                    & HI & 0.1907 & 0.0689 & 0.1880 & 0.1008 \\
                    & PA & 0.1968 & 0.1573 & 0.1940 & \textbf{0.1737} \\
\midrule
Zephyr-7B & EN & 0.1920 & 0.0467 & 0.1920 & 0.0751 \\
          & HI & 0.2028 & 0.0406 & 0.2000 & 0.0675 \\
          & PA & 0.2028 & 0.0406 & 0.2000 & 0.0675 \\
\bottomrule
\end{tabular}
\end{table}

These findings are specific to zero-shot closed-label prompting: the evaluated LLMs must map multilingual clinical notes directly to one of six predefined categories without task-specific adaptation. The results do not imply that fine-tuned or clinically adapted LLMs would exhibit the same behavior. In the present study, however, the supervised domain-adaptive encoder provides a substantially stronger diagnostic backbone for the subsequent deterministic selective verification analysis.

\begin{figure}[H]
  \centering
  \includegraphics[width=\linewidth]{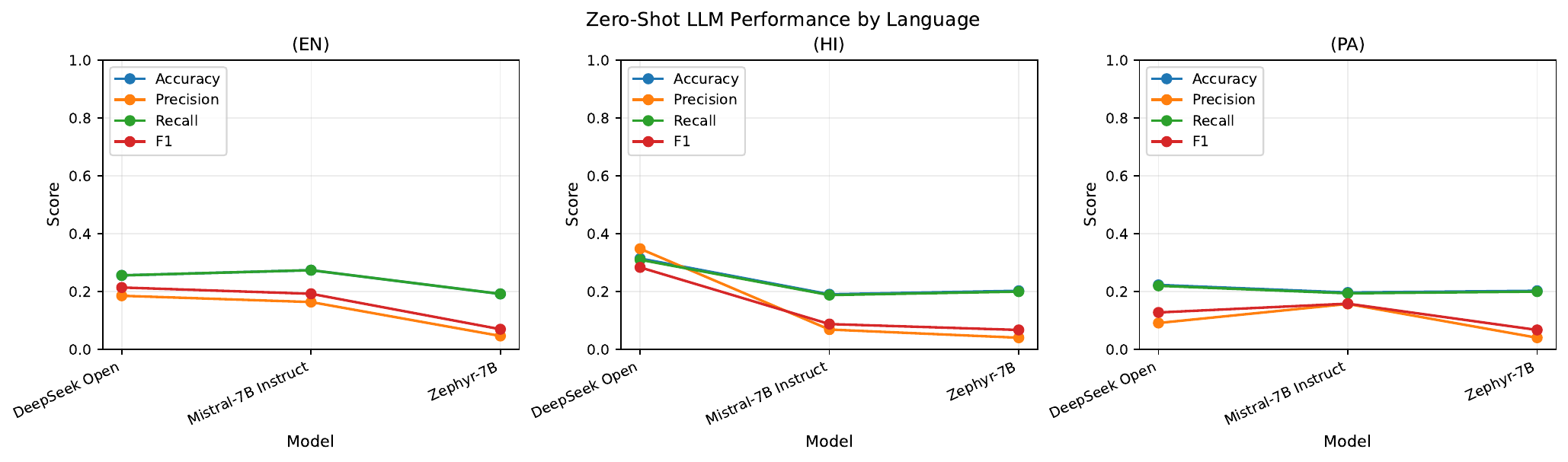}
  \caption{Zero-shot LLM performance across English, Hindi, and Punjabi under the controlled setting.}
  \label{fig:llm_zeroshot}
\end{figure}

\subsection{Natural-Prevalence Clinical Distribution}
\label{rcd}

The natural-prevalence setting evaluates model behavior under the diagnostic-label distribution observed in the refined clinical corpus. Compared with the controlled setting, this evaluation introduces substantial class imbalance and language-specific prevalence differences, making minority and ambiguous categories more challenging to identify reliably. We therefore examine class-wise performance before reporting aggregate natural-prevalence results.

\subsubsection{Unknown: Insufficient or Ambiguous Evidence}

The \textit{Unknown} category captures notes with insufficient or ambiguous evidence for a specific orthopedic category and remains distinct from the post-prediction \textit{defer} action. Under natural prevalence, IndicBERT-HPA achieves the highest F1-score in English, Hindi and Punjabi (0.887, 0.908 and 0.935), together with the highest ROC-AUC values (0.934, 0.935 and 0.951). Table~\ref{tab:unknown_uc} therefore shows consistent recognition of insufficient-evidence cases across all three languages.

\begin{table}[H]
  \caption{Prediction performance for \textbf{Unknown} under the natural-prevalence setting.}
  \label{tab:unknown_uc}
  \centering
  \renewcommand{\arraystretch}{0.80}
  \resizebox{\columnwidth}{!}{%
  \begin{tabular}{lcccc}
    \toprule
    \textbf{Model} &
    \textbf{Precision $\pm$ Std} &
    \textbf{Recall $\pm$ Std} &
    \textbf{F1} &
    \textbf{ROC-AUC $\pm$ Std} \\
    \midrule
     &
     \textbf{EN / HI / PA} &
     \textbf{EN / HI / PA} &
     \textbf{EN / HI / PA} &
     \textbf{EN / HI / PA} \\
    \midrule

    DistilBERT &
    0.636$\pm$0.020 / 0.597$\pm$0.020 / 0.695$\pm$0.020 &
    0.714$\pm$0.020 / 0.549$\pm$0.020 / 0.611$\pm$0.020 &
    0.673 / 0.572 / 0.650 &
    0.740$\pm$0.020 / 0.689$\pm$0.020 / 0.732$\pm$0.020 \\

    XLM-RoBERTa &
    0.758$\pm$0.020 / 0.719$\pm$0.020 / 0.809$\pm$0.020 &
    0.752$\pm$0.020 / 0.799$\pm$0.020 / 0.750$\pm$0.020 &
    0.755 / 0.757 / 0.778 &
    0.784$\pm$0.020 / 0.872$\pm$0.020 / 0.791$\pm$0.020 \\

    IndicBERT &
    0.810$\pm$0.020 / 0.586$\pm$0.020 / 0.676$\pm$0.020 &
    0.775$\pm$0.020 / 0.527$\pm$0.020 / 0.528$\pm$0.020 &
    0.792 / 0.555 / 0.593 &
    0.910$\pm$0.020 / 0.894$\pm$0.020 / 0.902$\pm$0.020 \\

    mDeBERTa &
    \textbf{0.904$\pm$0.020} / 0.871$\pm$0.020 / 0.794$\pm$0.020 &
    0.657$\pm$0.020 / 0.797$\pm$0.020 / 0.747$\pm$0.020 &
    0.761 / 0.832 / 0.770 &
    0.892$\pm$0.020 / 0.891$\pm$0.020 / 0.897$\pm$0.020 \\

    IndicBERT-HPA (Proposed) &
    0.893$\pm$0.020 / \textbf{0.923$\pm$0.020} / \textbf{0.949$\pm$0.020} &
    \textbf{0.881$\pm$0.020} / \textbf{0.893$\pm$0.020} / \textbf{0.922$\pm$0.020} &
    \textbf{0.887} / \textbf{0.908} / \textbf{0.935} &
    \textbf{0.934$\pm$0.020} / \textbf{0.935$\pm$0.020} / \textbf{0.951$\pm$0.020} \\

    \bottomrule
  \end{tabular}
  }
\end{table}

\subsubsection{Bone-Related Disorders}

Bone-related disorders provide consistent evidence for the effectiveness of IndicBERT-HPA under natural prevalence. The proposed model achieves the highest F1-score in English, Hindi and Punjabi (0.840, 0.842 and 0.884), while also recording the highest ROC-AUC in Hindi (0.890) and Punjabi (0.872). DistilBERT retains the highest English ROC-AUC (0.901). As reported in Table~\ref{tab:bone_related_disorders_uc}, these findings indicate strong thresholded performance across languages, with particularly strong class separation in the Hindi and Punjabi partitions.

\begin{table}[H]
  \caption{Prediction performance for \textbf{Bone-related disorders} under the natural-prevalence setting.}
  \label{tab:bone_related_disorders_uc}
  \centering
  \renewcommand{\arraystretch}{0.80}
  \resizebox{\columnwidth}{!}{%
  \begin{tabular}{lcccc}
    \toprule
    \textbf{Model} &
    \textbf{Precision $\pm$ Std} &
    \textbf{Recall $\pm$ Std} &
    \textbf{F1} &
    \textbf{ROC-AUC $\pm$ Std} \\
    \midrule
     &
     \textbf{EN / HI / PA} &
     \textbf{EN / HI / PA} &
     \textbf{EN / HI / PA} &
     \textbf{EN / HI / PA} \\
    \midrule

    DistilBERT &
    \textbf{0.822$\pm$0.020} / 0.792$\pm$0.020 / 0.710$\pm$0.020 &
    0.766$\pm$0.020 / 0.607$\pm$0.020 / 0.744$\pm$0.020 &
    0.793 / 0.687 / 0.727 &
    \textbf{0.901$\pm$0.020} / 0.841$\pm$0.020 / 0.824$\pm$0.020 \\

    XLM-RoBERTa &
    0.672$\pm$0.020 / 0.731$\pm$0.020 / 0.596$\pm$0.020 &
    0.587$\pm$0.020 / 0.697$\pm$0.020 / 0.721$\pm$0.020 &
    0.627 / 0.714 / 0.653 &
    0.481$\pm$0.020 / 0.577$\pm$0.020 / 0.531$\pm$0.020 \\

    IndicBERT &
    0.783$\pm$0.020 / 0.758$\pm$0.020 / 0.730$\pm$0.020 &
    0.755$\pm$0.020 / 0.770$\pm$0.020 / 0.843$\pm$0.020 &
    0.769 / 0.764 / 0.782 &
    0.797$\pm$0.020 / 0.821$\pm$0.020 / 0.714$\pm$0.020 \\

    mDeBERTa &
    0.461$\pm$0.020 / 0.434$\pm$0.020 / 0.420$\pm$0.020 &
    0.480$\pm$0.020 / 0.572$\pm$0.020 / 0.536$\pm$0.020 &
    0.470 / 0.494 / 0.471 &
    0.473$\pm$0.020 / 0.440$\pm$0.020 / 0.372$\pm$0.020 \\

    IndicBERT-HPA (Proposed) &
    0.813$\pm$0.020 / \textbf{0.855$\pm$0.020} / \textbf{0.873$\pm$0.020} &
    \textbf{0.869$\pm$0.020} / \textbf{0.830$\pm$0.020} / \textbf{0.896$\pm$0.020} &
    \textbf{0.840} / \textbf{0.842} / \textbf{0.884} &
    0.816$\pm$0.020 / \textbf{0.890$\pm$0.020} / \textbf{0.872$\pm$0.020} \\

    \bottomrule
  \end{tabular}
  }
\end{table}

\subsubsection{Hip-Related Disorders}

Hip-related classification remains strong for IndicBERT-HPA under natural prevalence, with the highest F1-score in English, Hindi and Punjabi (0.885, 0.889 and 0.865). The proposed model also achieves the highest ROC-AUC in Hindi (0.871) and Punjabi (0.892), whereas XLM-RoBERTa leads English ROC-AUC (0.900). The comparisons in Table~\ref{tab:hip_related_disorders} indicate consistently strong thresholded performance for IndicBERT-HPA, while class-separation behavior remains language dependent.

\begin{table}[H]
  \caption{Prediction performance for \textbf{Hip-related disorders} under the natural-prevalence setting.}
  \label{tab:hip_related_disorders}
  \centering
  \renewcommand{\arraystretch}{0.80}
  \resizebox{\columnwidth}{!}{%
  \begin{tabular}{lcccc}
    \toprule
    \textbf{Model} &
    \textbf{Precision $\pm$ Std} &
    \textbf{Recall $\pm$ Std} &
    \textbf{F1} &
    \textbf{ROC-AUC $\pm$ Std} \\
    \midrule
     &
     \textbf{EN / HI / PA} &
     \textbf{EN / HI / PA} &
     \textbf{EN / HI / PA} &
     \textbf{EN / HI / PA} \\
    \midrule

    DistilBERT &
    0.682$\pm$0.020 / 0.719$\pm$0.020 / 0.625$\pm$0.020 &
    0.669$\pm$0.020 / 0.741$\pm$0.020 / 0.707$\pm$0.020 &
    0.675 / 0.730 / 0.663 &
    0.809$\pm$0.020 / 0.750$\pm$0.020 / 0.752$\pm$0.020 \\

    XLM-RoBERTa &
    0.857$\pm$0.020 / 0.848$\pm$0.020 / 0.861$\pm$0.020 &
    0.562$\pm$0.020 / 0.873$\pm$0.020 / 0.753$\pm$0.020 &
    0.679 / 0.860 / 0.803 &
    \textbf{0.900$\pm$0.020} / 0.867$\pm$0.020 / 0.850$\pm$0.020 \\

    IndicBERT &
    0.814$\pm$0.020 / 0.652$\pm$0.020 / 0.720$\pm$0.020 &
    0.781$\pm$0.020 / 0.601$\pm$0.020 / 0.611$\pm$0.020 &
    0.797 / 0.625 / 0.661 &
    0.742$\pm$0.020 / 0.769$\pm$0.020 / 0.697$\pm$0.020 \\

    mDeBERTa &
    \textbf{0.899$\pm$0.020} / 0.825$\pm$0.020 / 0.771$\pm$0.020 &
    0.812$\pm$0.020 / 0.839$\pm$0.020 / 0.769$\pm$0.020 &
    0.853 / 0.832 / 0.770 &
    0.886$\pm$0.020 / 0.814$\pm$0.020 / 0.796$\pm$0.020 \\

    IndicBERT-HPA (Proposed) &
    0.893$\pm$0.020 / \textbf{0.893$\pm$0.020} / \textbf{0.875$\pm$0.020} &
    \textbf{0.877$\pm$0.020} / \textbf{0.886$\pm$0.020} / \textbf{0.856$\pm$0.020} &
    \textbf{0.885} / \textbf{0.889} / \textbf{0.865} &
    0.833$\pm$0.020 / \textbf{0.871$\pm$0.020} / \textbf{0.892$\pm$0.020} \\

    \bottomrule
  \end{tabular}
  }
\end{table}

\subsubsection{Musculoskeletal Disorders}

Musculoskeletal-disorder classification shows a consistent advantage for IndicBERT-HPA under natural prevalence. The proposed model achieves the highest F1-score in English, Hindi and Punjabi (0.886, 0.869 and 0.876), together with the highest ROC-AUC in Hindi (0.923) and Punjabi (0.937). IndicBERT records the highest English ROC-AUC (0.864), indicating that class-separation gains remain language dependent despite the stronger thresholded performance of IndicBERT-HPA across languages (Table~\ref{tab:musculoskeletal_disorders}).

\begin{table}[H]
  \caption{Prediction performance for \textbf{Musculoskeletal disorders} under the natural-prevalence setting.}
  \label{tab:musculoskeletal_disorders}
  \centering
  \renewcommand{\arraystretch}{0.80}
  \resizebox{\columnwidth}{!}{%
  \begin{tabular}{lcccc}
    \toprule
    \textbf{Model} &
    \textbf{Precision $\pm$ Std} &
    \textbf{Recall $\pm$ Std} &
    \textbf{F1} &
    \textbf{ROC-AUC $\pm$ Std} \\
    \midrule
     &
     \textbf{EN / HI / PA} &
     \textbf{EN / HI / PA} &
     \textbf{EN / HI / PA} &
     \textbf{EN / HI / PA} \\
    \midrule

    DistilBERT &
    0.808$\pm$0.020 / 0.764$\pm$0.020 / 0.675$\pm$0.020 &
    0.767$\pm$0.020 / 0.617$\pm$0.020 / 0.523$\pm$0.020 &
    0.787 / 0.683 / 0.589 &
    0.817$\pm$0.020 / 0.751$\pm$0.020 / 0.761$\pm$0.020 \\

    XLM-RoBERTa &
    0.692$\pm$0.020 / 0.692$\pm$0.020 / 0.729$\pm$0.020 &
    0.834$\pm$0.020 / 0.834$\pm$0.020 / 0.736$\pm$0.020 &
    0.756 / 0.756 / 0.732 &
    0.822$\pm$0.020 / 0.794$\pm$0.020 / 0.806$\pm$0.020 \\

    IndicBERT &
    0.813$\pm$0.020 / 0.675$\pm$0.020 / 0.704$\pm$0.020 &
    0.757$\pm$0.020 / 0.667$\pm$0.020 / 0.731$\pm$0.020 &
    0.784 / 0.671 / 0.717 &
    \textbf{0.864$\pm$0.020} / 0.838$\pm$0.020 / 0.877$\pm$0.020 \\

    mDeBERTa &
    0.826$\pm$0.020 / 0.784$\pm$0.020 / 0.803$\pm$0.020 &
    0.785$\pm$0.020 / 0.726$\pm$0.020 / 0.881$\pm$0.020 &
    0.805 / 0.754 / 0.840 &
    0.835$\pm$0.020 / 0.848$\pm$0.020 / 0.902$\pm$0.020 \\

    IndicBERT-HPA (Proposed) &
    \textbf{0.894$\pm$0.020} / \textbf{0.829$\pm$0.020} / \textbf{0.863$\pm$0.020} &
    \textbf{0.879$\pm$0.020} / \textbf{0.913$\pm$0.020} / \textbf{0.889$\pm$0.020} &
    \textbf{0.886} / \textbf{0.869} / \textbf{0.876} &
    0.813$\pm$0.020 / \textbf{0.923$\pm$0.020} / \textbf{0.937$\pm$0.020} \\

    \bottomrule
  \end{tabular}
  }
\end{table}

\subsubsection{Spinal Disorders}

Spinal-disorder classification further demonstrates the strong thresholded performance of IndicBERT-HPA under natural prevalence. The proposed model achieves the highest F1-score in English, Hindi and Punjabi (0.904, 0.908 and 0.866), together with the highest ROC-AUC in Hindi (0.925) and Punjabi (0.915). In English, mDeBERTa records the highest ROC-AUC (0.903), indicating metric-specific competition in this partition. Overall, Table~\ref{tab:spinal_disorders} shows that IndicBERT-HPA maintains strong performance across languages with particularly effective class separation in Hindi and Punjabi.

\begin{table}[H]
  \caption{Prediction performance for \textbf{Spinal disorders} under the natural-prevalence setting.}
  \label{tab:spinal_disorders}
  \centering
  \renewcommand{\arraystretch}{0.80}
  \resizebox{\columnwidth}{!}{%
  \begin{tabular}{lcccc}
    \toprule
    \textbf{Model} &
    \textbf{Precision $\pm$ Std} &
    \textbf{Recall $\pm$ Std} &
    \textbf{F1} &
    \textbf{ROC-AUC $\pm$ Std} \\
    \midrule
     &
     \textbf{EN / HI / PA} &
     \textbf{EN / HI / PA} &
     \textbf{EN / HI / PA} &
     \textbf{EN / HI / PA} \\
    \midrule

    DistilBERT &
    0.740$\pm$0.020 / 0.766$\pm$0.020 / 0.706$\pm$0.020 &
    0.727$\pm$0.020 / 0.648$\pm$0.020 / 0.637$\pm$0.020 &
    0.733 / 0.702 / 0.670 &
    0.851$\pm$0.020 / 0.847$\pm$0.020 / 0.750$\pm$0.020 \\

    XLM-RoBERTa &
    0.711$\pm$0.020 / 0.871$\pm$0.020 / 0.795$\pm$0.020 &
    0.719$\pm$0.020 / 0.869$\pm$0.020 / 0.818$\pm$0.020 &
    0.715 / 0.870 / 0.806 &
    0.884$\pm$0.020 / 0.820$\pm$0.020 / 0.808$\pm$0.020 \\

    IndicBERT &
    0.817$\pm$0.020 / 0.781$\pm$0.020 / 0.724$\pm$0.020 &
    0.865$\pm$0.020 / 0.746$\pm$0.020 / 0.807$\pm$0.020 &
    0.840 / 0.763 / 0.763 &
    0.760$\pm$0.020 / 0.832$\pm$0.020 / 0.864$\pm$0.020 \\

    mDeBERTa &
    0.736$\pm$0.020 / 0.794$\pm$0.020 / 0.815$\pm$0.020 &
    0.746$\pm$0.020 / 0.793$\pm$0.020 / 0.827$\pm$0.020 &
    0.741 / 0.793 / 0.821 &
    \textbf{0.903$\pm$0.020} / 0.870$\pm$0.020 / 0.812$\pm$0.020 \\

    IndicBERT-HPA (Proposed) &
    \textbf{0.923$\pm$0.020} / \textbf{0.893$\pm$0.020} / \textbf{0.875$\pm$0.020} &
    \textbf{0.886$\pm$0.020} / \textbf{0.923$\pm$0.020} / \textbf{0.857$\pm$0.020} &
    \textbf{0.904} / \textbf{0.908} / \textbf{0.866} &
    0.875$\pm$0.020 / \textbf{0.925$\pm$0.020} / \textbf{0.915$\pm$0.020} \\

    \bottomrule
  \end{tabular}
  }
\end{table}

\subsubsection{Other (Heterogeneous Category)}

The heterogeneous \textit{Other} category exhibits substantial metric-dependent variation across languages under natural prevalence. IndicBERT-HPA achieves the highest F1-score in Hindi and Punjabi (0.909 and 0.909) and the highest Punjabi ROC-AUC (0.925), whereas XLM-RoBERTa leads English F1-score (0.866) and Hindi ROC-AUC (0.910), and mDeBERTa records the highest English ROC-AUC (0.976). As detailed in Table~\ref{tab:other}, no single model dominates every language--metric comparison in this category, although IndicBERT-HPA remains strongest for the Hindi and Punjabi fixed-threshold results.

\begin{table}[H]
  \caption{Prediction performance for \textbf{Other} under the natural-prevalence setting.}
  \label{tab:other}
  \centering
  \renewcommand{\arraystretch}{0.80}
  \resizebox{\columnwidth}{!}{%
  \begin{tabular}{lcccc}
    \toprule
    \textbf{Model} &
    \textbf{Precision $\pm$ Std} &
    \textbf{Recall $\pm$ Std} &
    \textbf{F1} &
    \textbf{ROC-AUC $\pm$ Std} \\
    \midrule
     &
     \textbf{EN / HI / PA} &
     \textbf{EN / HI / PA} &
     \textbf{EN / HI / PA} &
     \textbf{EN / HI / PA} \\
    \midrule

    DistilBERT &
    0.604$\pm$0.020 / 0.639$\pm$0.020 / 0.599$\pm$0.020 &
    0.729$\pm$0.020 / 0.643$\pm$0.020 / 0.539$\pm$0.020 &
    0.661 / 0.641 / 0.567 &
    0.799$\pm$0.020 / 0.774$\pm$0.020 / 0.748$\pm$0.020 \\

    XLM-RoBERTa &
    \textbf{0.926$\pm$0.020} / 0.845$\pm$0.020 / 0.831$\pm$0.020 &
    0.814$\pm$0.020 / 0.733$\pm$0.020 / 0.780$\pm$0.020 &
    \textbf{0.866} / 0.785 / 0.805 &
    0.852$\pm$0.020 / \textbf{0.910$\pm$0.020} / 0.836$\pm$0.020 \\

    IndicBERT &
    0.704$\pm$0.020 / 0.676$\pm$0.020 / 0.669$\pm$0.020 &
    0.872$\pm$0.020 / 0.636$\pm$0.020 / 0.780$\pm$0.020 &
    0.779 / 0.655 / 0.720 &
    0.934$\pm$0.020 / 0.854$\pm$0.020 / 0.895$\pm$0.020 \\

    mDeBERTa &
    0.723$\pm$0.020 / 0.898$\pm$0.020 / 0.821$\pm$0.020 &
    \textbf{0.900$\pm$0.020} / 0.860$\pm$0.020 / 0.665$\pm$0.020 &
    0.802 / 0.879 / 0.735 &
    \textbf{0.976$\pm$0.020} / 0.842$\pm$0.020 / 0.873$\pm$0.020 \\

    IndicBERT-HPA (Proposed) &
    0.700$\pm$0.020 / \textbf{0.924$\pm$0.020} / \textbf{0.841$\pm$0.020} &
    0.832$\pm$0.020 / \textbf{0.895$\pm$0.020} / \textbf{0.990$\pm$0.020} &
    0.760 / \textbf{0.909} / \textbf{0.909} &
    0.902$\pm$0.020 / 0.891$\pm$0.020 / \textbf{0.925$\pm$0.020} \\

    \bottomrule
  \end{tabular}
  }
\end{table}

\FloatBarrier
Under natural prevalence, IndicBERT-HPA consistently leads all language partitions in F1-Macro, Macro-AUROC, AUPRC and accuracy, as reported in Tables~\ref{tab:transformer_performance_languages} and~\ref{tab:transformer_average_performance}. Its F1-Macro reaches 0.8605, 0.8876 and 0.8894, with corresponding accuracies of 0.873, 0.885 and 0.880 for English, Hindi and Punjabi, respectively. These results indicate stable classification performance despite language-specific prevalence differences. Among the baselines, XLM-RoBERTa attains the highest averaged F1-Macro (0.7621), IndicBERT the highest Macro-AUROC (0.831), and mDeBERTa the highest AUPRC (0.800).

\begin{table*}[t]
\centering
\caption{Task-aligned encoder performance by language under the natural-prevalence setting. Macro-AUROC denotes the unweighted mean of the one-vs-rest ROC-AUC values across the six diagnostic categories.}
\label{tab:transformer_performance_languages}
\scriptsize
\setlength{\tabcolsep}{2pt}
\renewcommand{\arraystretch}{0.80}
\begin{tabular}{lcccccc}
\toprule
\textbf{Model} & \textbf{Lang.} & \textbf{F1-Macro} & \textbf{Macro-AUROC} & \textbf{AUPRC} & \textbf{ECE} & \textbf{Accuracy} \\
\midrule
DistilBERT & EN & 0.7204 & 0.820 & 0.725 & 0.083 & 0.708 \\
           & HI & 0.6691 & 0.775 & 0.693 & 0.089 & 0.684 \\
           & PA & 0.6445 & 0.761 & 0.706 & 0.118 & 0.630 \\
\midrule
XLM-RoBERTa & EN & 0.7330 & 0.787 & 0.761 & 0.083 & 0.647 \\
            & HI & 0.7904 & 0.807 & 0.748 & 0.103 & 0.827 \\
            & PA & 0.7630 & 0.770 & 0.742 & 0.103 & 0.785 \\
\midrule
IndicBERT & EN & 0.7936 & 0.834 & 0.769 & 0.072 & 0.783 \\
          & HI & 0.6723 & 0.835 & 0.739 & 0.099 & 0.641 \\
          & PA & 0.7062 & 0.825 & 0.747 & \textbf{0.074} & 0.782 \\
\midrule
mDeBERTa & EN & 0.7387 & 0.828 & 0.806 & 0.074 & 0.764 \\
         & HI & 0.7640 & 0.784 & 0.810 & 0.089 & 0.782 \\
         & PA & 0.7345 & 0.775 & 0.783 & 0.092 & 0.775 \\
\midrule
IndicBERT-HPA (Proposed)
          & EN & \textbf{0.8605} & \textbf{0.862} & \textbf{0.921} & \textbf{0.049} & \textbf{0.873} \\
          & HI & \textbf{0.8876} & \textbf{0.906} & \textbf{0.903} & \textbf{0.077} & \textbf{0.885} \\
          & PA & \textbf{0.8894} & \textbf{0.915} & \textbf{0.882} & 0.104 & \textbf{0.880} \\
\bottomrule
\end{tabular}
\end{table*}

When averaged across languages, IndicBERT-HPA obtains 0.8792 F1-Macro, 0.894 Macro-AUROC, 0.902 AUPRC and 0.879 accuracy, outperforming all evaluated baselines overall (Table~\ref{tab:transformer_average_performance}). It also achieves the lowest averaged ECE (0.077); however, its higher Punjabi ECE relative to several baselines indicates that calibration remains language dependent and motivates deterministic selective verification.

\begin{table}[t]
\centering
\caption{Average encoder performance across languages under the natural-prevalence setting. Macro-AUROC is averaged across diagnostic categories and subsequently summarized across languages.}
\label{tab:transformer_average_performance}
\scriptsize
\setlength{\tabcolsep}{2pt}
\renewcommand{\arraystretch}{0.85}
\begin{tabular}{lccccc}
\toprule
\textbf{Model} & \textbf{F1-Macro} & \textbf{Macro-AUROC} & \textbf{AUPRC} & \textbf{ECE} & \textbf{Accuracy} \\
\midrule
DistilBERT               & 0.6780 & 0.785 & 0.708 & 0.097 & 0.674 \\
XLM-RoBERTa              & 0.7621 & 0.788 & 0.750 & 0.096 & 0.753 \\
IndicBERT                & 0.7240 & 0.831 & 0.752 & 0.082 & 0.736 \\
mDeBERTa                 & 0.7457 & 0.796 & 0.800 & 0.085 & 0.774 \\
IndicBERT-HPA (Proposed) & \textbf{0.8792} & \textbf{0.894} & \textbf{0.902} & \textbf{0.077} & \textbf{0.879} \\
\bottomrule
\end{tabular}
\end{table}

\subsection{Zero-Shot Evaluation of Large Language Models Under Natural Prevalence}
\label{sec:zeroshot}

The three zero-shot LLMs are evaluated under the natural-prevalence setting using the protocol defined in Section~\ref{subsec:llms}. DeepSeek Open achieves the highest averaged F1-score (0.6121), followed by Zephyr-7B (0.5994) and Mistral-7B Instruct (0.4790). It also leads in English and Hindi, whereas Zephyr-7B obtains the highest Punjabi F1-score (0.5848). Although zero-shot performance improves relative to the controlled setting, all three LLMs remain substantially below IndicBERT-HPA, which achieves an averaged F1-Macro of 0.8792 under the same natural-prevalence test setting (Table~\ref{tab:llm_performance}).

\begin{table*}[t]
\centering
\caption{Zero-shot LLM performance under the natural-prevalence setting; All reports the average across languages.}
\label{tab:llm_performance}

\scriptsize
\setlength{\tabcolsep}{2pt}
\renewcommand{\arraystretch}{0.85}

\begin{tabular}{lccccc}
\toprule
\textbf{Model} & \textbf{Lang.} & \textbf{Accuracy} & \textbf{Precision} & \textbf{Recall} & \textbf{F1-Score} \\
\midrule
DeepSeek Open & EN & 0.6554 & 0.6712 & 0.6934 & \textbf{0.6821} \\
              & HI & 0.6523 & 0.5836 & 0.6572 & \textbf{0.6182} \\
              & PA & 0.5561 & 0.5045 & 0.5718 & 0.5361 \\
\midrule
Mistral-7B Instruct & EN & 0.5382 & 0.4869 & 0.5648 & 0.5230 \\
                    & HI & 0.4947 & 0.4155 & 0.5494 & 0.4732 \\
                    & PA & 0.4735 & 0.4378 & 0.4437 & 0.4407 \\
\midrule
Zephyr-7B & EN & 0.5828 & 0.6130 & 0.6113 & 0.6122 \\
          & HI & 0.5839 & 0.5987 & 0.6036 & 0.6011 \\
          & PA & 0.5905 & 0.5597 & 0.6122 & \textbf{0.5848} \\
\midrule
DeepSeek Open      & All & \textbf{0.6213} & 0.5864 & \textbf{0.6408} & \textbf{0.6121} \\
Mistral-7B Instruct & All & 0.5021 & 0.4467 & 0.5193 & 0.4790 \\
Zephyr-7B           & All & 0.5857 & \textbf{0.5905} & 0.6090 & 0.5994 \\
\bottomrule
\end{tabular}
\end{table*}

Zero-shot LLM performance remains language dependent under natural prevalence: all three models obtain lower Punjabi than English F1-scores, and the strongest model varies across language partitions. These findings are limited to the evaluated zero-shot closed-label setting and do not characterize fine-tuned or clinically adapted LLM systems. Within the present study, the supervised domain-adaptive encoder therefore provides the stronger classification backbone for deterministic selective verification.

\begin{figure*}[t]
    \centering
    \captionsetup[subfigure]{skip=0pt}

    \begin{subfigure}[t]{0.31\textwidth}
        \centering
        \includegraphics[width=\linewidth]{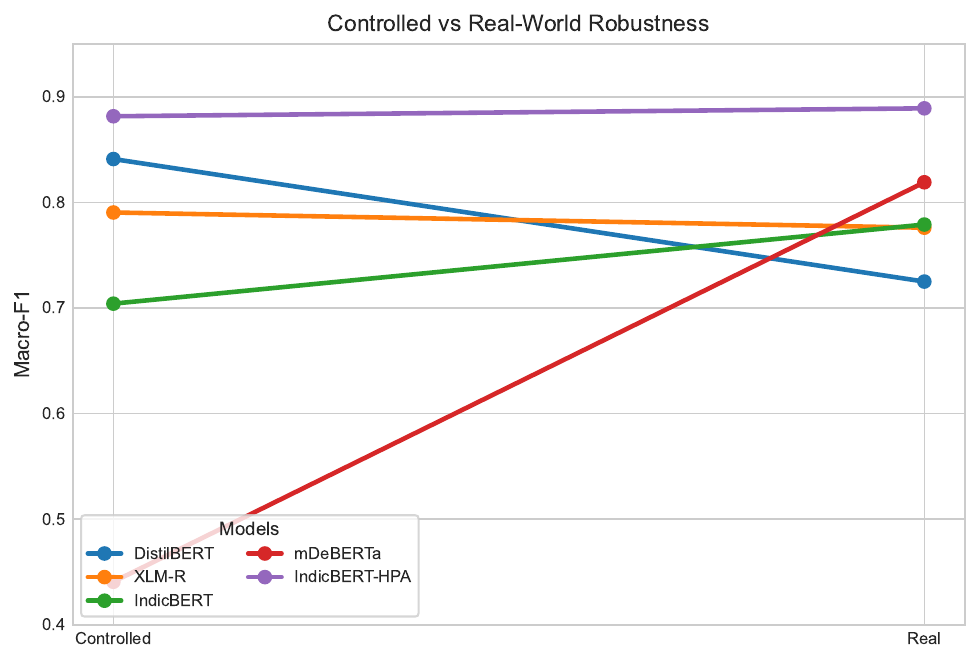}
        \caption{Controlled vs.\ natural prevalence.}
        \label{fig:robustness_slope}
    \end{subfigure}
    \hfill
    \begin{subfigure}[t]{0.31\textwidth}
        \centering
        \includegraphics[width=\linewidth]{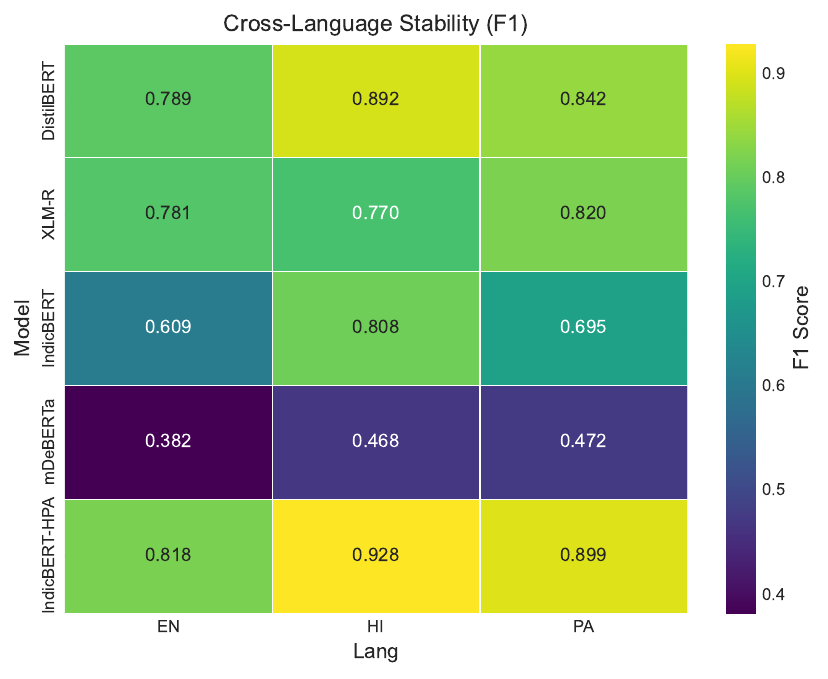}
        \caption{Language-wise performance.}
        \label{fig:language_heatmap}
    \end{subfigure}
    \hfill
    \begin{subfigure}[t]{0.31\textwidth}
        \centering
        \includegraphics[width=\linewidth]{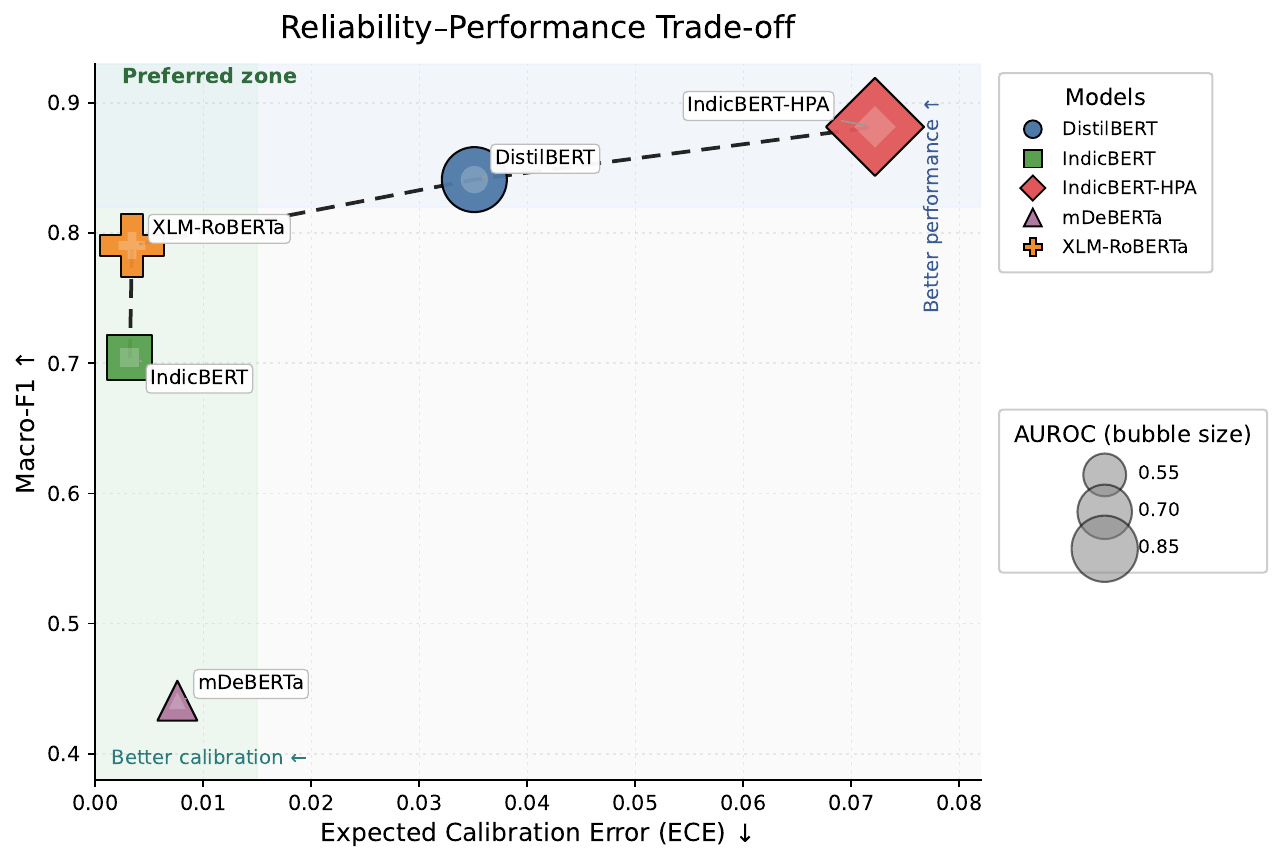}
        \caption{Calibration and performance.}
        \label{fig:reliability_tradeoff}
    \end{subfigure}

    \vspace{-0.45em}

    \begin{subfigure}[t]{0.31\textwidth}
        \centering
        \includegraphics[width=\linewidth]{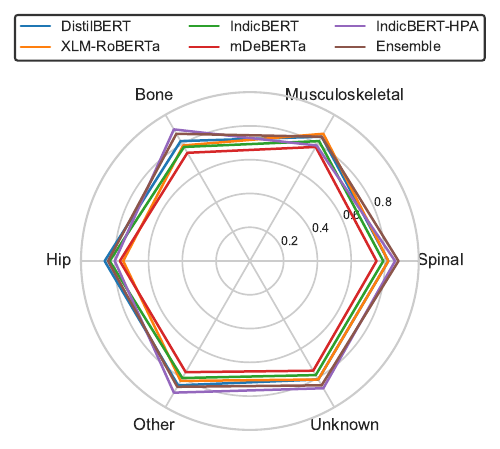}
        \caption{Class-wise performance.}
        \label{fig:class_radar}
    \end{subfigure}
    \hfill
    \begin{subfigure}[t]{0.31\textwidth}
        \centering
        \includegraphics[width=\linewidth]{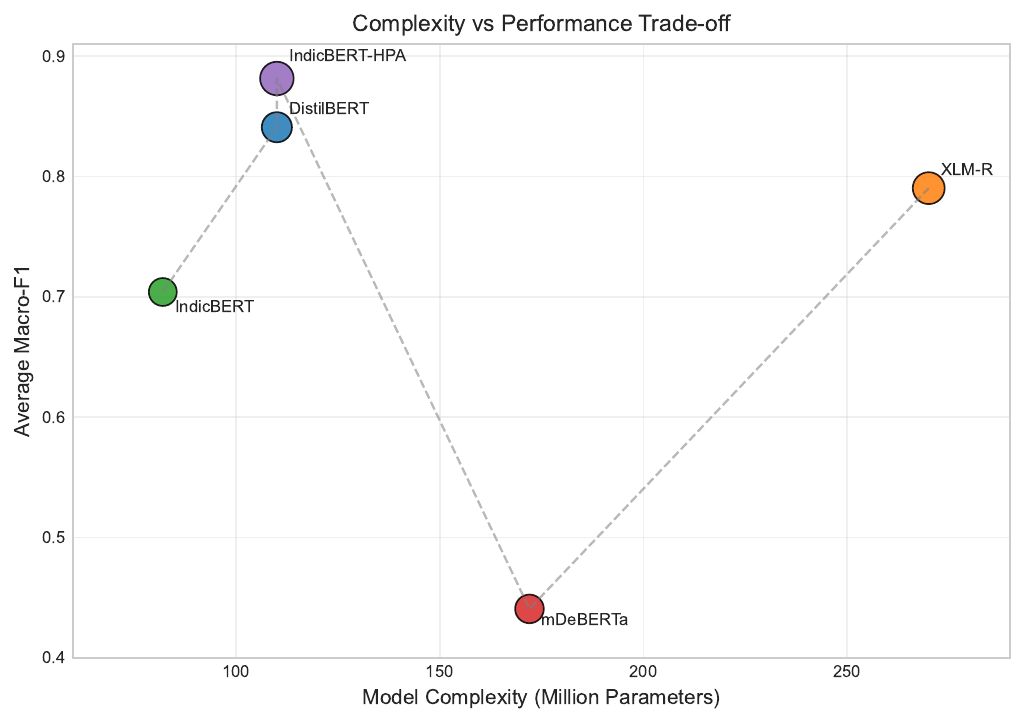}
        \caption{Complexity and performance.}
        \label{fig:complexity_tradeoff}
    \end{subfigure}
    \hfill
    \begin{subfigure}[t]{0.31\textwidth}
        \centering
        \includegraphics[width=\linewidth]{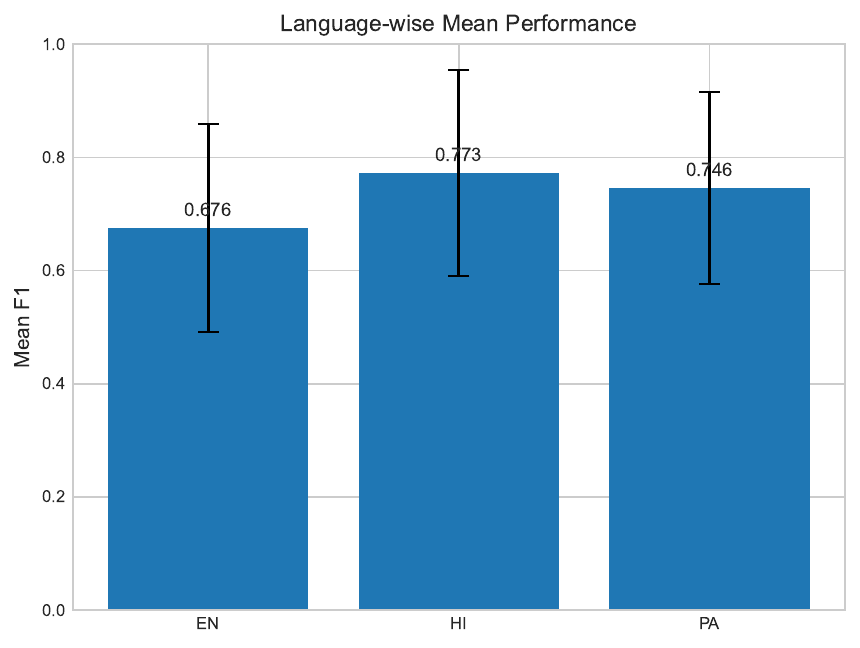}
        \caption{Mean performance by language.}
        \label{fig:language_mean}
    \end{subfigure}

   \caption{Complementary model analyses under controlled and natural-prevalence settings.}
    \label{fig:combined_visualizations}
\end{figure*}

Figure~\ref{fig:combined_visualizations} summarizes distributional, language-wise, calibration, class-wise and complexity-related trade-offs, reinforcing the separate evaluation of predictive performance and confidence reliability for post-prediction verification.

\subsection{Deterministic Selective Verification Analysis}
\label{sec:agent_results}

The controlled and natural-prevalence results motivate a post-prediction reliability analysis. Zero-shot LLMs remain weaker than supervised encoders for closed-label multilingual classification, whereas IndicBERT-HPA provides the strongest discrimination but still shows language-dependent calibration variation. We therefore retrospectively evaluate the deterministic selective-verification layer defined in Section~\ref{sec:verification}. Without modifying the predicted category, the layer combines confidence gating, symptom--diagnosis evidence checking and language-consistency screening to accept or defer each prediction.

\subsection{Retrospective Selective-Verification Evaluation on a Randomly Selected Held-Out Subset}
\label{sec:verification_validation}

We retrospectively evaluate the deterministic selective-verification layer on a randomly selected subset of 5,000 records from the held-out natural-prevalence test partition, comprising 1,644 English, 1,682 Hindi and 1,674 Punjabi notes. Applied to fixed IndicBERT-HPA predictions, the layer uses confidence, symptom--diagnosis evidence and language-risk signals to authorize automatic acceptance or defer a case for review, without modifying the classifier or its predicted category.

This subset is evaluated separately from the complete natural-prevalence test evaluation reported in Tables~\ref{tab:transformer_performance_languages} and~\ref{tab:transformer_average_performance}. All accept-all and selective-verification comparisons reported in this section are computed on the same randomly selected 5,000-record subset. Therefore, the subset-specific accept-all baseline should be interpreted only as the reference point for the selective accuracy--coverage analysis in this section and not as a replacement for the aggregate full-test performance reported earlier.

\subsubsection{Subset-Specific Baseline Performance Without Verification}

Before selective verification, IndicBERT-HPA is evaluated on all 5,000 records in the held-out verification subset using an accept-all policy. It achieves 71.5\% overall accuracy and 0.65 Macro-F1, with lower performance in Punjabi than in English and Hindi. These results provide the subset-specific baseline for assessing whether deterministic deferral improves the reliability of automatically accepted predictions (Table~\ref{tab:base_without_verification}).

\begin{table}[t]
\centering
\caption{Subset-specific accept-all performance of IndicBERT-HPA on the randomly selected 5,000-record held-out natural-prevalence verification subset.}
\label{tab:base_without_verification}
\footnotesize
\setlength{\tabcolsep}{8pt}
\renewcommand{\arraystretch}{1.05}
\begin{tabular}{lcc}
\toprule
\textbf{Language} & \textbf{Accuracy} & \textbf{Macro-F1} \\
\midrule
English & 73.2\% & 0.68 \\
Hindi   & 71.8\% & 0.66 \\
Punjabi & 69.5\% & 0.62 \\
\midrule
\textbf{Overall} & \textbf{71.5\%} & \textbf{0.65} \\
\bottomrule
\end{tabular}
\end{table}

\subsubsection{Cross-Language Selective Performance}

Applied to the same held-out subset, the complete verification layer accepts 72.3\% of predictions and defers 27.7\% for review. Among accepted outputs, it achieves 84.4\% selective accuracy and 0.76 selective Macro-F1, with selective accuracy remaining above 83\% in all three languages. These results indicate that the policy identifies a more reliable accepted subset across languages (Table~\ref{tab:cross_language_verification}). The increase from 71.5\% accept-all accuracy to 84.4\% selective accuracy should be interpreted jointly with coverage: it does not indicate improved accuracy over all 5,000 records. Instead, the layer reduces errors among automatically accepted predictions by deferring cases with insufficient confidence, unsupported evidence or elevated language-related risk.

\begin{table}[t]
\centering
\caption{Cross-language selective performance of the complete verification layer on the randomly selected 5,000-record held-out natural-prevalence verification subset.}
\label{tab:cross_language_verification}
\footnotesize
\setlength{\tabcolsep}{5pt}
\renewcommand{\arraystretch}{1.05}
\begin{tabular}{lcccc}
\toprule
\textbf{Language} & \textbf{Sel.-Acc.} & \textbf{Sel.-F1} & \textbf{Coverage} & \textbf{Deferral} \\
\midrule
English & 85.4\% & 0.78 & 75.0\% & 25.0\% \\
Hindi   & 84.6\% & 0.76 & 72.0\% & 28.0\% \\
Punjabi & 83.2\% & 0.74 & 70.0\% & 30.0\% \\
\midrule
\textbf{Overall} & \textbf{84.4\%} & \textbf{0.76} & \textbf{72.3\%} & \textbf{27.7\%} \\
\bottomrule
\end{tabular}
\end{table}

\subsubsection{Operating-Policy Comparison}

Table~\ref{tab:verification_policy_comparison} reports representative operating points for independently configured policies on the same randomly selected held-out subset. The accept-all reference retains every prediction, whereas the remaining policies condition automatic acceptance on confidence, evidence, language risk or their combination.

Because these policies operate at different coverage levels, Table~\ref{tab:verification_policy_comparison} should not be interpreted as a cumulative component-wise ablation or as isolating individual verification signals. Instead, it reports the selective accuracy--coverage trade-off across reliability-control policies, with the complete verification layer achieving the highest selective accuracy (84.4\%) at 72.3\% coverage.

\begin{table}[t]
\centering
\caption{Representative operating points of independently configured selective-verification policies on the randomly selected held-out natural-prevalence verification subset. These policies are not cumulative component-wise ablations.}
\label{tab:verification_policy_comparison}

\scriptsize
\setlength{\tabcolsep}{4pt}
\renewcommand{\arraystretch}{0.95}

\begin{tabular}{lccc}
\toprule
\textbf{Policy} & \textbf{Selective Accuracy} & \textbf{Coverage} & \textbf{Deferral Rate} \\
\midrule
Accept-all baseline         & 71.5\%          & 100.0\% & 0.0\%  \\
Confidence-gated policy     & 80.2\%          & 75.0\%  & 25.0\% \\
Evidence-aware policy       & 82.0\%          & 72.0\%  & 28.0\% \\
Language-risk policy        & 83.6\%          & 70.0\%  & 30.0\% \\
Complete verification layer & \textbf{84.4\%} & 72.3\%  & 27.7\% \\
\bottomrule
\end{tabular}
\end{table}

\subsection{Selective Prediction and Illustrative Review Scenario}
\label{subsec:illustrative_review_scenario}

Table~\ref{tab:selective_safety} summarizes the subset-specific selective-verification results. Relative to the accept-all reference, which yields 71.5\% accuracy, 0.65 Macro-F1 and approximately 1,425 incorrect predictions, the complete verification layer accepts 72.3\% of cases with 84.4\% selective accuracy and 0.76 selective Macro-F1. It leaves approximately 564 incorrect outputs in the accepted stream and routes approximately 861 for review, corresponding to a 60.4\% reduction in unsafe accepted errors through deferral rather than diagnostic correction. Under the illustrative, non-validated assumption that review corrects 70\% of routed errors, approximately 603 errors could potentially be corrected. These results are retrospective and subset-specific and do not establish population-level performance, clinician performance or workflow-level benefit.

\begin{table}[t]
\centering
\caption{Selective-prediction outcomes and an illustrative review-correction scenario on the randomly selected held-out natural-prevalence verification subset; the scenario is hypothetical and not clinician validated.}
\label{tab:selective_safety}

\scriptsize
\setlength{\tabcolsep}{4pt}
\renewcommand{\arraystretch}{0.95}

\begin{tabular}{lc}
\toprule
\textbf{Metric} & \textbf{Value} \\
\midrule
\multicolumn{2}{l}{\textit{Empirical selective-prediction behavior}} \\
Coverage / auto-accept rate & 72.3\% \\
Deferral / review rate & 27.7\% \\
Selective accuracy among accepted cases & 84.4\% \\
Overall selective Macro-F1 among accepted cases & 0.76 \\
Accepted predictions, approximately & 3,615 \\
Deferred predictions, approximately & 1,385 \\
Accept-all incorrect predictions, approximately & 1,425 \\
Incorrect accepted outputs remaining, approximately & 564 \\
Incorrect outputs routed for review, approximately & 861 \\
Unsafe accepted error reduction relative to accept-all & 60.4\% \\
\midrule
\multicolumn{2}{l}{\textit{Hypothetical review-correction scenario}} \\
Assumed correction effectiveness & 70.0\% \\
Potentially corrected routed errors, approximately & 603 \\
\bottomrule
\end{tabular}
\end{table}

\paragraph{Summary of Selective Verification Results.}
On the randomly selected held-out subset, the verification layer accepts 72.3\% of predictions and achieves 84.4\% selective accuracy and 0.76 selective Macro-F1, compared with 71.5\% accuracy and 0.65 Macro-F1 under accept-all prediction. Its 60.4\% reduction in unsafe accepted errors demonstrates an accuracy--coverage trade-off through selective deferral. These findings are subset-specific retrospective evidence and do not establish population-level performance or prospective clinical workflow benefit.

\section{Discussion}
\label{sec:discussion}

The results show that multilingual orthopedic text classification should be evaluated beyond aggregate predictive performance. IndicBERT-HPA achieves the strongest averaged controlled-setting profile across F1-Macro, Macro-AUROC, AUPRC and accuracy, and remains the strongest encoder under natural prevalence, reaching 0.8792 F1-Macro, 0.894 Macro-AUROC, 0.902 AUPRC and 0.879 accuracy with the lowest averaged ECE of 0.077. Its comparatively higher Punjabi ECE nevertheless shows that language-aware orthopedic adaptation improves classification quality without eliminating language-dependent calibration risk.

The zero-shot results distinguish general instruction-following capability from task-aligned clinical classification. Under natural prevalence, even the strongest evaluated LLM, DeepSeek Open, reaches an averaged F1-score of 0.6121, compared with 0.8792 for IndicBERT-HPA; DeepSeek Open leads in English and Hindi, whereas Zephyr-7B leads in Punjabi. These findings apply only to zero-shot closed-label prompting without clinical fine-tuning or task-specific adaptation and should not be generalized to clinically adapted LLM systems.

On the randomly selected 5,000-record held-out verification subset, accept-all IndicBERT-HPA achieves 71.5\% accuracy and 0.65 Macro-F1, whereas the deterministic selective-verification layer accepts 72.3\% of predictions with 84.4\% selective accuracy and 0.76 selective Macro-F1, deferring 27.7\% for review. Based on rounded values, this corresponds to a 60.4\% reduction in unsafe accepted errors. This result reflects an accuracy--coverage trade-off produced by selective deferral rather than improved classification over all cases. The subset serves as a retrospective reliability stress test, not a replacement for the complete natural-prevalence evaluation; prospective clinician-in-the-loop validation remains necessary before workflow-level use.

\subsection{Limitations}
\label{subsec:limitations}

This study has four principal limitations. First, the deterministic selective-verification layer is evaluated retrospectively on a randomly selected held-out subset rather than across the complete natural-prevalence test partition or in a prospective clinical workflow. All selective analyses use the same 5,000-record subset, whose accept-all baseline differs from aggregate full-test performance; therefore, the reported coverage, deferral, selective accuracy, selective Macro-F1 and unsafe accepted error reduction constitute subset-specific evidence rather than population- or workflow-level findings. Future validation should extend this evaluation to the full test partition, repeated random subsets and prospective clinician-in-the-loop assessment of safety, usability, review burden, clinician agreement and effects on clinical decision-making.

Second, the study is limited to orthopedic narratives in English, Hindi and Punjabi, and the natural-prevalence partitions exhibit substantial language-conditioned label skew: hip-related cases dominate the English and Hindi partitions, whereas spinal cases dominate the Punjabi partition. Consequently, observed differences may reflect linguistic, prevalence, documentation or collection effects. Evaluation with matched language--label distributions, broader language and specialty coverage, and external institutional cohorts is required to assess generalizability.

Third, the LLM comparison is restricted to three instruction-tuned models evaluated through zero-shot closed-label prompting. No task-specific fine-tuning, retrieval augmentation, parameter-efficient adaptation, or clinically adapted LLM configuration is assessed. Moreover, the reported LLM analysis focuses on classification performance rather than formal calibration or selective-prediction evaluation. Conclusions regarding LLM behavior should therefore be interpreted only within the evaluated zero-shot setting and should not be generalized to clinically adapted or task-fine-tuned LLM systems.

Finally, the de-identified clinical records cannot be publicly released because of patient privacy, institutional governance and ethical-approval constraints. Although the preprocessing, training, evaluation and deterministic verification procedures are specified, complete external replication requires controlled access to suitable de-identified data and verification resources. Future work should develop governed access procedures and privacy-preserving, synthetic or de-identified evaluation resources for broader methodological verification without compromising confidentiality.

\section{Conclusion and Future Work}
\label{sec:conclusion}

This study investigated reliability-oriented multilingual orthopedic text classification across English, Hindi and Punjabi clinical narratives. IndicBERT-HPA, which combines a shared IndicBERT backbone with language-aware orthopedic adapters, achieves the strongest natural-prevalence performance, with averaged Macro-F1 of 0.8792, Macro-AUROC of 0.894, AUPRC of 0.902 and accuracy of 0.879. In contrast, the strongest evaluated zero-shot LLM reaches an averaged F1-score of 0.6121, underscoring the importance of task-aligned adaptation in this setting. On a randomly selected 5,000-record held-out subset, deterministic selective verification accepts 72.3\% of predictions and achieves 84.4\% selective accuracy and 0.76 selective Macro-F1, compared with 71.5\% accuracy and 0.65 Macro-F1 under accept-all prediction, while reducing unsafe accepted errors by approximately 60.4\% through deferral. These findings support domain-adaptive multilingual encoders with deterministic post-prediction verification, while prospective clinician-in-the-loop evaluation and external validation remain necessary before workflow-level use.

\paragraph{Future work.}
Future work will evaluate deterministic selective verification over the complete natural-prevalence test partition, repeated random subsets and prospective clinician-in-the-loop settings. We will also compare task-adapted and retrieval-supported LLMs with IndicBERT-HPA under the same reliability metrics, and extend external evaluation to additional clinical specialties and low-resource languages to assess generalizability, language--label distribution effects and calibration stability.

\section*{Data Availability}
The dataset underlying the results presented in this study contains sensitive clinical information. Due to privacy and IRB constraints, it can be made available to qualified researchers upon reasonable request to the authors.

\section*{Code Availability}
To preserve double-anonymous review, the identifying public repository link is omitted from the review manuscript. The implementation and experimental configuration files will be made available in the non-anonymous version of the article.

\section*{Ethical Consideration}
This study was conducted in full compliance with ethical standards for research involving human clinical data. The study protocol and data usage were reviewed and approved by the Institutional Review Board (IRB) at hospital (anonymous), approval number anonymous, ensuring adherence to ethical guidelines, patient privacy and informed consent practices. All patient data were fully anonymized prior to sharing with removal of personally identifiable information to preserve confidentiality and comply with privacy regulations.

\bibliographystyle{ACM-Reference-Format}
\bibliography{Reference}

\end{document}